\pdfoutput=1

\documentclass[11pt]{article}

\usepackage[]{acl}

\usepackage{times}
\usepackage{latexsym}

\usepackage[T1]{fontenc}

\usepackage[utf8]{inputenc}

\usepackage{microtype}

\usepackage{booktabs}
\usepackage{graphicx}
\usepackage{multirow}
\usepackage{subcaption}
\usepackage{color, colortbl}
\usepackage{graphics}
\usepackage{amsmath}
\usepackage{lipsum}
\usepackage{algorithm}
\usepackage{algpseudocode}
\usepackage{comment}
\usepackage{makecell}
\usepackage{bm}
\usepackage{amssymb}
	
\definecolor{LightBlue}{rgb}{0.80,0.93,1}
\definecolor{LightGreen}{rgb}{0.88,0.97,0.84}
\definecolor{LightGray}{rgb}{0.91,0.91,0.91}

\title{Interactive Concept Learning for Uncovering Latent Themes
in Large Text Collections}

\author{
   Maria Leonor Pacheco$^{1,2}$ \,\,
   Tunazzina Islam$^{3}$ \,\,
   Lyle Ungar$^4$ \,\,
   Ming Yin$^3$ \,\,
   Dan Goldwasser$^3$ \\
    $^1$Microsoft Research~~~
    $^2$University of Colorado Boulder \\
    $^3$Purdue University~~~
    $^4$University of Pennsylvania \\
    \texttt{maria.pacheco@colorado.edu}~~ \texttt{ungar@cis.upenn.edu}\\
    \texttt{\{islam32,mingyin,dgoldwas\}@purdue.edu} 
}

\begin{document}
\maketitle
\begin{abstract}
Experts across diverse disciplines are often interested in making sense of large text collections. Traditionally, this challenge is approached either by noisy unsupervised techniques such as topic models, or by following a manual theme discovery process. In this paper, we expand the definition of a theme to account for more than just a word distribution, and include generalized concepts deemed relevant by domain experts. Then, we propose an interactive framework that receives 
 and encodes expert feedback at different levels of abstraction. Our framework strikes a balance between automation and manual coding, allowing experts to maintain control of their study while reducing the manual effort required. 
\end{abstract}

\section{Introduction}

Researchers and practitioners across diverse disciplines are often interested making sense of large text collections. Thematic analysis is one of the most common qualitative research methods used to approach this challenge, and it can be understood as a form of pattern recognition in which the themes (or codes) that emerge from the data become the categories for analysis \cite{thematic_analysis,Roberts2019AttemptingRA}. In standard practice, researchers bring their own objectives or questions and identify the relevant themes or patterns recognized while analyzing the data, potentially grounding them in a relevant theory or framework. Themes in thematic analysis are broadly defined as ``patterned responses or meaning'' derived from the data, which inform the research question.

With the explosion of data and the rapid development of automated techniques, disciplines that traditionally relied on qualitative methods for the analysis of textual content are turning to computational methods~\cite{doi:10.1146/annurev-polisci-090216-023229,IJoC10675}. Topic modeling has long been the go-to NLP technique to identify emerging themes from text collections \cite{10.5555/944919.944937,8186620,doi:10.1080/19312458.2021.2015574}. Despite its wide adoption, topic modeling does not afford the same flexibility and representation power of qualitative techniques. For this reason, many efforts have been dedicated to understanding the ways in which topic models can be flawed~\cite{mimno-etal-2011-optimizing}, and evaluating their coherence and quality~\cite{stevens-etal-2012-exploring,lau-etal-2014-machine,10.1145/2684822.2685324}. More recently, ~\citet{NEURIPS2021_0f83556a} showed that human judgements and accepted metrics of topic quality and coherence do not always agree. Given the noisy landscape surrounding topic modeling, qualitative methods are still prevalent across fields for analyzing nuanced and verbally complex data~\cite{rose_and_lennerholt,8490223,antons_et_al}. 

Human-in-the-loop topic modeling approaches aim to address these issues by allowing experts to correct and influence the output of topic models. Given that topics in topic models are defined as distributions over words, the feedback received using these approaches is usually limited to identifying representative words and imposing constraints between words~\cite{hu-etal-2011-interactive,lund-etal-2017-tandem,10.1145/3172944.3172965}. In this paper, we argue that themes emerging from a document collection should not just be defined as a word distribution (similar to a topic model), but as a distribution over generalized concepts that can help us explain them. We build on the definition put forward by \citet{thematic_analysis}, where themes are latent patterned meanings that emerge from the data, and supporting concepts serve as a way to explain themes using theoretical frameworks that are deemed relevant by domain experts. For example, emerging themes in a dataset about Covid-19 can be characterized by the strength of their relationship to stances about the covid vaccine and the moral framing of relevant entities (e.g. The theme \textit{``Government distrust''} is strongly correlated to an \textit{anti-vax} stance and frames \textit{Dr. Fauci} as an entity enabling \textit{cheating}). This representation of a theme aligns more closely with qualitative practices, as experts can introduce their pre-existing knowledge about the domain. Moreover, higher-level abstractions expand the capabilities of experts to correct and influence theme discovery, as it allows them to formulate concepts to generalize from observations to new examples \cite{Rogers2004SemanticCA}, and to deductively draw inferences via conceptual rules and statements \cite{johnson_1988}. 

Following this rationale, we suggest a new computational approach to support and enhance standard qualitative practices for content analysis. We approach both inductive thematic analysis (i.e. identifying the relevant themes that emerge from the data and developing the code-book), and deductive thematic analysis (i.e. identifying the instances where a discovered theme is observed). To support this process, we allow researchers to shape the space of themes given machine generated candidates. Then, we allow them to provide feedback over machine judgments that map text to themes using relevant conceptual frameworks. 

To showcase our approach, we look at the task of characterizing social media discussions around topics of interest to the computational social science community. Namely, we consider two distinct case studies: The covid-19 vaccine debate in the US, and the immigration debate in the US, the UK and Europe. For each case, the qualitative researchers use different theories to ground theme discovery, each associated with a different set of concepts. For the covid-19 vaccine debate, the theme discovery process is grounded using vaccination stances and morality frames \cite{roy-etal-2021-identifying,pacheco-etal-2022-holistic}. For the immigration debates, the theme discovery is grounded using three framing typologies: narrative frames \cite{alma991011943249704551}, policy frames \cite{card-etal-2015-media} and immigration frames \cite{benson_2013,doi:10.1080/13183222.2019.1589285}. All of these choices build on previous work and were validated by the qualitative researchers. From a machine learning perspective, these two case studies could be regarded as completely different tasks and have been approached independently in previous work. The reason for this is the data, the context, and the target labels (both the emerging themes and the supporting concepts) are different for each scenario.


To aid experts in theme discovery, we propose an iterative two-stage machine-in-the-loop framework. In the first stage, we provide experts with an automated partition of the data, ranked example instances, and visualizations of the concept distribution. Then, we have a group of experts work together to explore the partitions, code emerging patterns and identify coherent themes. Once themes are identified, we have the experts select representative examples, write down additional examples and explanatory phrases, and explain themes using the set of available concepts. In the second stage, we incorporate the expert feedback using a neuro-symbolic mapping procedure. The \textit{symbolic} part allows us to explicitly model the dependencies between concepts and the emerging themes using weighted logical formulae (e.g. $w: \mathtt{policy\_frame(economic)} \Rightarrow \mathtt{theme(economic\_migrants)}$. These rules can be interpreted as soft constraints whose weights are learned from the feedback provided by the experts. The \textit{neural} part allows us to maintain a distributed representation of the data points and themes, which facilitates the live exploration of the data based on distances and similarities, and provides a feature representation for learning the rule weights. After the mapping stage concludes, some instances will be assigned to the identified themes, and the remaining instances will be re-partitioned for a consecutive discovery stage.

We conducted extensive evaluations of the different components, design choices, and stages in our methodology. We showed that our framework allows experts to uncover a set of themes that cover a large portion of the data, and that the resulting mapping from tweets to themes is fairly accurate with respect to human judgements. While we focused on polarized discussions, our framework generalizes to any content analysis study where the space of relevant themes is not known in advance.




\section{Related Work}
This paper suggests a novel approach for identifying themes emerging from text collections. The notion of a theme presented in this work is strongly related to topic models~\cite{10.5555/944919.944937}. However, unlike latent topics that are defined as word distributions, our goal is to provide a richer representation that more strongly resembles qualitative practices by connecting the themes to general concepts that help explain them. For example, when identifying themes emerging from polarized discussions in social media, we look at conceptual frameworks such as moral foundations theory~\cite{haidt2007morality,amin2017association,chan2021moral} and framing theory~\cite{entman1993framing,chong2007framing,morstatter2018identifying}.

Our work is conceptually similar to recent contributions that characterize themes and issue-specific frames in data, either by manually developing a code-book and annotating the data according to it~\cite{boydstun2014tracking,mendelsohn-etal-2021-modeling}, or by using data-driven methods~\cite{demszky2019analyzing,roy-goldwasser-2021-analysis}. Unlike these approaches, our work relies on interleaved human-machine interaction rounds, in which humans can identify and explain themes from a set of candidates suggested by the model, as well as diagnose and adapt the model's ability to recognize these themes in documents. This work is part of a growing trend in NLP that studies how human-machine collaboration can help improve automated language analysis~\cite{wang-etal-2021-putting}. In that space, two lines of works are most similar to ours. Interactive topic models~\cite{hu-etal-2011-interactive,lund-etal-2017-tandem,10.1145/3172944.3172965} allows humans to adapt the identified topics using lexical information. Open Framing~\cite{bhatia-etal-2021-openframing} allows humans to identify and name frames based on the output of topic models, but lacks our model's ability for sustained interactions that help shape the theme space, as well as the explanatory power of our neuro-symbolic representation. 


%

\section{The Framework}\label{sec:framework}

\begin{figure}[t]
    \centering
    \includegraphics[width=0.8\columnwidth]{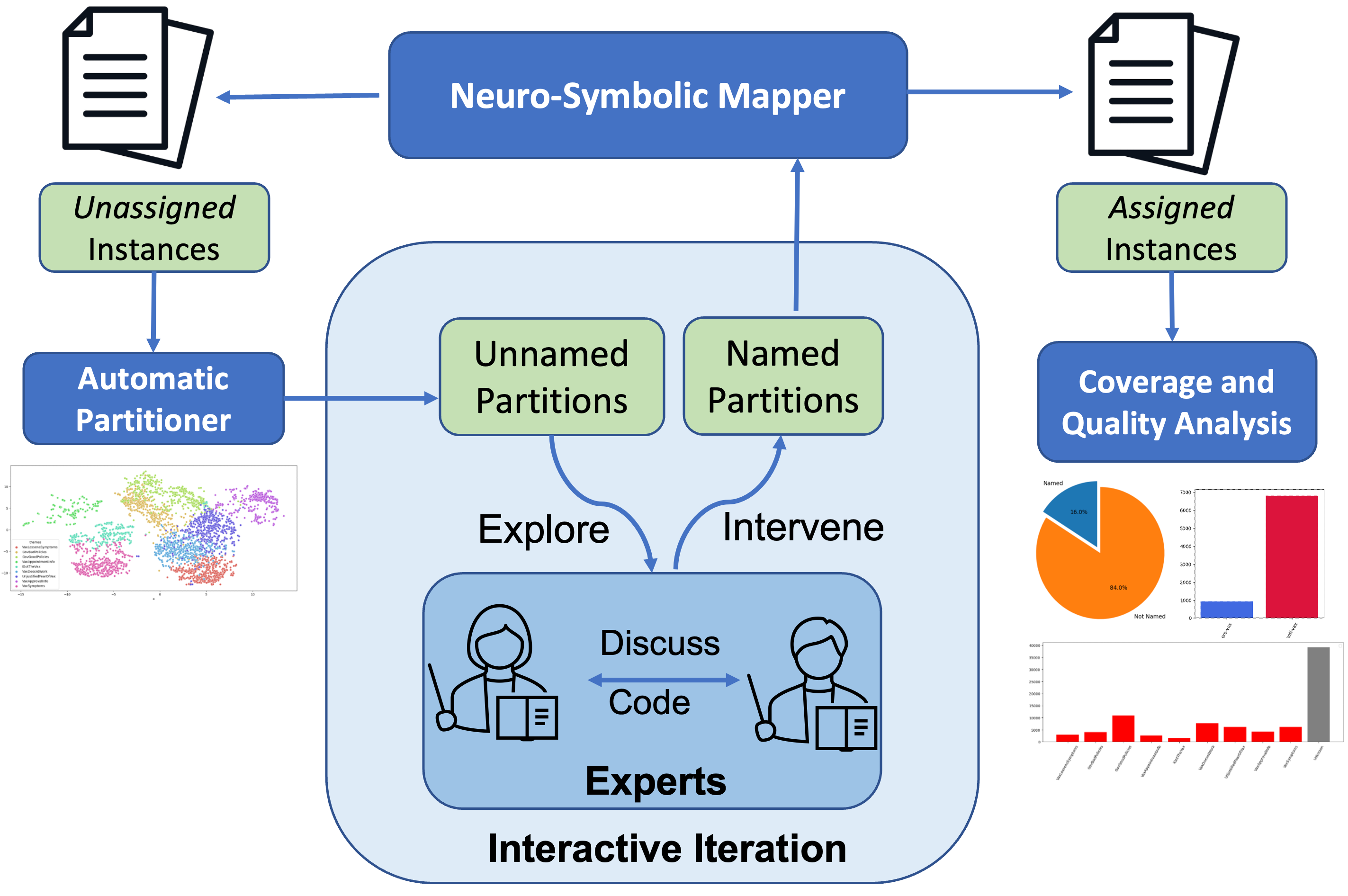}
    \caption{Framework Overview}
    \label{fig:framework}
\end{figure}

We propose an iterative two-stage framework that combines ML/NLP techniques, interactive interfaces and qualitative methods to assist experts in characterizing large textual collections. We define large textual collections as repositories of textual instances (e.g. tweets, posts, documents), where each instance is potentially associated with a set of annotated or predicted concepts. 

In the first stage, our framework automatically proposes an initial partition of the data, such that instances that are thematically similar are clustered together. We provide experts with an interactive interface equipped with a set of \textit{exploratory operations} that allows them to evaluate the quality of the discovered partitions, as well as to further explore and partition the space by inspecting individual examples, finding similar instances, and using open text queries. As experts interact with the data through the interface, they following an inductive thematic analysis approach to identify and code the patterns that emerge within the partitions~\cite{thematic_analysis}. Next, they group the identified patterns into general themes, and instantiate them using the interface. Although intuitively we could expect a single partition to result in a single theme, note that this is not enforced. Experts maintain full freedom as to how many themes they instantiate, if any. Once a theme is created, experts are provided with a set of \textit{intervention operations} to explain the themes using natural language, select good example instances, write down additional examples, and input or correct supporting concepts to characterize the theme assignments. The full set of operations are listed in Tab. \ref{tab:ops} and demonstrated in App.~\ref{app:tool}.



In the second stage, our framework finds a mapping between the full set of instances and the themes instantiated by the experts. We use the information contributed by the experts in the form of examples and concepts, and learn to map instances to themes using our neuro-symbolic procedure. We allow instances to remain unassigned if there is not a good enough match, and in this case, a consecutive portioning step is done. We refer to instances that are mapped to themes as ``named partitions'' and unassigned proposed partitions as ``unnamed partitions''. Once instances are assigned to themes, experts have access to a comprehensive visual analysis of the state of the system. The main goal of this analysis is to appreciate the trade-off between coverage (how many instances we can account for with the discovered themes) and quality (how good we are at mapping instances to themes). An illustration of the framework can be observed in Fig. \ref{fig:framework}. Additional details about the coverage and quality analysis are presented in the experimental section.

Below, we discuss the representation of themes and instances, the protocol followed for interaction, and the mapping and re-partitioning procedures.


\paragraph{Representing Themes and Instances} We represent example instances and explanatory phrases using their S-BERT embedding \cite{reimers-gurevych-2019-sentence}. To measure the closeness between an instance and a theme, we compute the cosine similarity between the instance and all of the explanatory phrases and examples for the theme, and take the maximum similarity score among them. Our framework is agnostic of the representation used. The underlying embedding objective, as well as the scoring function can easily be replaced. 

\begin{table}[ht]
    \centering
    \begin{subtable}{\columnwidth}
     \scalebox{0.60}{\begin{tabular}{|>{\arraybackslash}m{2cm}|>{\arraybackslash}m{9cm}|}
     \hline
       \textbf{Operations}  &  \textbf{Description} \\
       \hline
       Finding Partitions  &  Experts can find partitions in the space of unassigned instances. We currently support the K-means~\cite{Jin2010} and Hierarchical Density-Based Clustering~\cite{mcinnes2017hdbscan} algorithms. 
       \\
       \hline
       Text-based Queries & Experts can type any query in natural language and find instances that are close to the query in the embedding space. \\
       \hline
       Finding Similar Instances & Experts have the ability to select each instance and find other examples that are close in the embedding space.\\      
       \hline
       Listing Themes and Instances  & Experts can browse the current list of themes and their mapped instances. Instances are ranked in order of ``goodness'', corresponding to the similarity in the embedding space to the theme representation. They can be listed from closest to most distant, or from most distant to closest.  \\
       \hline
       Visualizing Local Explanations & Experts can visualize aggregated statistics and explanations for each of the themes. To obtain these explanations, we aggregate all instances that have been identified as being associated with a theme. Explanations include wordclouds, frequent entities and their sentiments, and graphs of concept distributions. \\
       \hline
       Visualizing Global Explanations & Experts can visualize aggregated statistics and explanations for the global state of the system. To do this, we aggregate all instances in the database. Explanations include theme distribution, coverage statistics, and t-sne plots \cite{vanDerMaaten2008}. \\
       \hline
       
    \end{tabular}}
    \caption{Exploratory Operations}\label{tab:exploratory_ops}
    \end{subtable}
    \begin{subtable}{\columnwidth}
\scalebox{0.60}{\begin{tabular}{|>{\arraybackslash}m{2cm}|>{\arraybackslash}m{9cm}|}
     \hline
       \textbf{Operations}  &  \textbf{Description} \\
       \hline
       Adding, Editing and Removing Themes & Experts can create, edit, and remove themes. The only requirement for creating a new theme is to give it a unique name. Similarly, themes can be edited or removed at any point. If any instances are assigned to a theme being removed, they will be moved to the space of unassigned instances.  \\
       \hline
       Adding and Removing Examples  & Experts can assign ``good'' and ``bad'' examples to existing themes. Good examples are instances that characterize the named theme. Bad examples are instances that could have similar wording to a good example, but that have different meaning. Experts can add examples in two ways: they can mark mapped instances as ``good'' or ``bad'', or they can directly contribute example phrases. \\
       \hline
       Adding or Correcting Concepts & We allow users to upload additional observed or predicted concepts for each textual instance. For instances and phrases added as ``good'' and ``bad'' examples, we allow users to add or edit the values of these concepts. The intuition behind this operation is to collect additional information for learning to map instances to themes. \\
       \hline
    \end{tabular}}
    \caption{Intervention Operations}\label{tab:intervention_ops}
    \end{subtable}
    \caption{Interactive Operations}\label{tab:ops}
\end{table}

\paragraph{Interaction Protocol} We follow a simple protocol where three human coders work together using the operations described above to discover themes in large textual corpora. In addition to the three coders, each interactive session is guided by one of the authors of the paper, who makes sure the coders are adhering to the process outlined here. 

To initialize the system, the coders will start by using the partitioning operation to find ten initial partitions of roughly the same size. During the first session, the coders will inspect the partitions one by one by looking at the examples closest to the centroid. This will be followed by a discussion phase, in which the coders follow an inductive thematic analysis approach to identify repeating patterns and write them down. If one or more cohesive patterns are identified, the experts will create a new theme, name it, and mark a set of good example instances that help in characterizing the named theme. When a pattern is not obvious, coders will explore similar instances to the different statements found. Whenever the similarity search results in a new pattern, the coders will create a new theme, name it, and mark a set of good example instances that helped in characterizing the named theme. 

Next, the coders will look at the local theme explanations and have the option to enhance each theme with additional phrases. Note that each theme already contains a small set of representative instances, which are marked as ``good'' in the previous step. In addition to contributing ``good'' example phrases, coders will have the option to contribute some ``bad'' example phrases to push the representation of the theme away from statements that have high lexical overlap with the good examples, but different meaning. Finally, coders will examine each exemplary instance and phrase for the set of symbolic concepts (e.g. stance, moral frames). In cases where the judgement is perceived as wrong, the coders will be allowed to correct it. In this paper, we assume that the textual corpora include a set of relevant concepts for each instance. In future work, we would like to explore the option of letting coders define concepts on the fly. 

\paragraph{Mapping and Re-partitioning} Each interactive session will be followed by a mapping and re-partitioning stage. First, we will perform the mapping step, in which we assign instances to the themes discovered during interaction. We do not assume that experts will have discovered the full space of latent themes. For this reason, we do not try to assign a theme to each and every instance. We expect that the set of themes introduced by the human experts at each round of interaction will cover a subset of the total instances available. Following this step, we will re-partition all the unassigned instances for a subsequent round of interaction. 





We use DRaiL~\cite{pacheco-goldwasser-2021-modeling}, a neuro-symbolic modeling framework to design a mapping procedure. Our main goal is to condition new theme assignments not only on the embedding distance between instances and good/bad examples, but also leverage the additional judgements provided by experts using the ``Adding or Correcting Concepts'' procedure. For example, when analyzing the corpus about the Covid-19 vaccine, experts could point out that 80\% of the good examples for theme \textit{Natural Immunity is Effective} have a clear \textit{anti-vaccine} stance. We could use this information to introduce inductive bias into our mapping procedure, and potentially capture cases where the embedding distance does not provide enough information. DRaiL uses weighted first-order logic rules to express decisions and dependencies between different decisions. We introduce the following rules:

\begin{equation*}\small
\begin{split}
t_0 - t_n: & \mathtt{Inst(i)} \Rightarrow \mathtt{Theme(i,t)} \\
a_0 - a_m: & \mathtt{Inst(i)} \Rightarrow  \mathtt{Concept(i,c)} \\
c_0 - c_{n*m}: & \mathtt{Inst(i)} \wedge \mathtt{Concept(i,c)} \Rightarrow \mathtt{Theme(i,t)} \\
c'_0 - c'_{n*n}: & \mathtt{Inst(i)} \wedge \mathtt{Theme(i,t)} \wedge (\mathtt{t} \neq \mathtt{t'}) \\
& \Rightarrow \mathtt{\neg Theme(i,t)}
\end{split}
\end{equation*}

The first set of rules $t_0 - t_n$ and $a_0 - a_m$ map instances to themes and concepts respectively. We create one template for each theme $\texttt{t}$ and concept $\texttt{c}$, and they correspond to binary decisions (e.g. whether instance $\texttt{i}$ mentions theme $\texttt{t}$). Then, we introduce two sets of soft constraints: $c_0 - c_{n*m}$ encode the dependencies between each concept and theme assignment (e.g. likelihood of theme \textit{Natural Immunity is Effective} given that instance has concept \textit{anti-vax}). Then, $c'_0 - c'_{n*n}$ discourages an instance from having more than one theme assignment. For each rule, we will learn a weight that captures the strength of that rule being active. Then, a combinatorial inference procedure will be run to find the most likely global assignment. Each entity and relation in DRaiL is tied to a neural architecture that is used to learn its weights. In this paper, we use a BERT encoder~\cite{devlin-etal-2019-bert} for all rules. To generate data for learning the DRaiL model, we take the $K=100$ closest instances for each good/bad example provided by the experts. Good examples will serve as positive training data. For negative training data, we take the contributed bad examples, as well as good examples for other themes and concepts. Once the weights are learned, we run the inference procedure over the full corpus. 

\section{Case Studies}


\begin{figure*}[ht]
    
    \begin{subfigure}{.25\linewidth}

         
    \includegraphics[width=\textwidth]{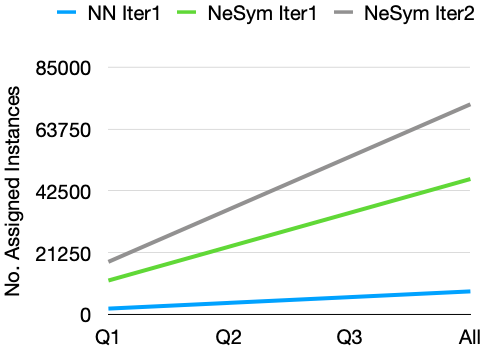}
    \caption{\textbf{Covid-19} Coverage}
    \label{tab:my_label}
    
    \end{subfigure}%
    \hfill
    \begin{subfigure}{.23\linewidth}
    \includegraphics[width=\textwidth]{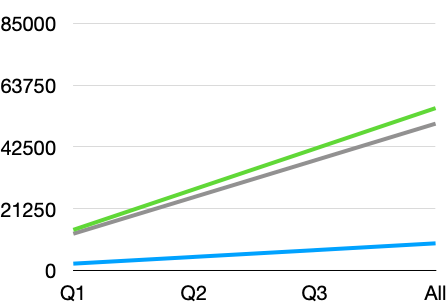}
    
    \caption{\textbf{Immigration} Coverage}
    \label{tab:my_label}
    
    \end{subfigure}
    \hfill
    \begin{subfigure}{.4\linewidth}
    
    \resizebox{\columnwidth}{!}{%
    \begin{tabular}{lllllll}
        \toprule
        Case & \multirow{2}{*}{Iter.} & Ground.  & \multirow{2}{*}{$\leq Q_1$} & \multirow{2}{*}{$\leq Q_2$} & \multirow{2}{*}{$\leq Q_3$} & \multirow{2}{*}{All}  \\
        Study &  & Method & & & &\\
        \midrule
        
        \multirow{3}{*}{\textbf{Covid-19}} & \cellcolor{LightBlue}1 &  \cellcolor{LightBlue}NNs & \cellcolor{LightBlue}89.80 & \cellcolor{LightBlue}87.50 & \cellcolor{LightBlue}87.50 & \cellcolor{LightBlue}85.71  \\
        
         & \cellcolor{LightGreen}  & \cellcolor{LightGreen} NeSym & \cellcolor{LightGreen} 87.50 & \cellcolor{LightGreen} 81.32 & \cellcolor{LightGreen} 75.38 & \cellcolor{LightGreen} 70.66  \\
        \cmidrule{2-7}

       & \cellcolor{LightGray} 2 & \cellcolor{LightGray} NeSym & \cellcolor{LightGray} 85.71 & \cellcolor{LightGray} 76.92 & \cellcolor{LightGray} 73.13 & \cellcolor{LightGray} 68.49 \\
        \midrule
       \multirow{3}{*}{\textbf{Immigration}} & \cellcolor{LightBlue}
        1 &  \cellcolor{LightBlue}NNs & \cellcolor{LightBlue}86.96 & \cellcolor{LightBlue}76.19 & \cellcolor{LightBlue}74.19 & \cellcolor{LightBlue}70.54  \\
        & \cellcolor{LightGreen}
         &  \cellcolor{LightGreen}NeSym & \cellcolor{LightGreen}85.29 & \cellcolor{LightGreen}79.07 & \cellcolor{LightGreen}73.51 & \cellcolor{LightGreen}70.54  \\
       \cmidrule{2-7}
       
        & \cellcolor{LightGray}
        2 & \cellcolor{LightGray}NeSym & \cellcolor{LightGray}91.43 & \cellcolor{LightGray}83.08 & \cellcolor{LightGray}79.15 & \cellcolor{LightGray}76.76 \\
        \bottomrule
    \end{tabular}
    }
    \caption{Theme F1}
    \label{tab:precision_covid}
    \end{subfigure}
   
    \caption{Theme Assignments Where Distance to Theme Centroid $\leq$ Quartile}\label{fig:general}
\end{figure*}

We explore two case studies involving discussions on social media: (1) The Covid-19 vaccine discourse in the US, and (2) The immigration discourse in the US, the UK and the EU. For the Covid-19 case, we build on the corpus of 85K tweets released by \citet{pacheco-etal-2022-holistic}. All tweets in this corpus were posted by users located in the US, are uniformly distributed between Jan. and Oct. 2021, and contain predictions for vaccination stance (e.g. pro-vax, anti-vax) and morality frames (e.g. fairness/cheating and their actor/targets.)~\cite{haidt2007morality, roy-etal-2021-identifying}. For the immigration case, we build on the corpus of 2.66M tweets released by \citet{mendelsohn-etal-2021-modeling}. All tweets in this corpus were posted by users located in the US, the UK and the EU, written between 2018 and 2019, and contain predictions for three different framing typologies: narrative frames (e.g. episodic, thematic)~\cite{alma991011943249704551}, generic policy frames (e.g. economic, security and defense, etc.)~\cite{card-etal-2015-media}, and immigration-specific frames (e.g. victim of war, victim of discrimination, etc.)~\cite{benson_2013,doi:10.1080/13183222.2019.1589285}. Additional details about the datasets and framing typologies can be found the original publications.

Our main goal is to evaluate whether experts can leverage our framework to identify prominent themes in the corpora introduced above. We recruited a group of six experts in Computational Social Science, four male and two female, within the ages of 25 and 45. The group of experts included advanced graduate students, postdoctoral researchers and faculty. Our studies are IRB approved, and we followed their protocols. For each corpus, we performed two consecutive sessions with three experts following the protocol outlined in Sec.~\ref{sec:framework}. To evaluate consistency, we did an additional two sessions with a different group of experts for the Covid-19 dataset.  Each session lasted a total of one hour. 
In App.~\ref{app:covid_sessions_first}, ~\ref{app:covid_sessions_second} and ~\ref{app:immi_sessions}, we include large tables enumerating the resulting themes, and describing in detail all of the patterns identified and coded by the experts at each step of the process.

\paragraph{Coverage vs. Mapping Quality} We evaluated the trade-off between coverage (how many tweets we can account for with the discovered themes) and mapping quality (how good we are at mapping tweets to themes). Results are outlined in Fig. \ref{fig:general}. To do this evaluation, we sub-sampled a set of 200 mapped tweets for each scenario, uniformly distributed across themes and their proximity to the theme embedding, and validated their assignments manually. The logic behind sampling across different proximities is that we expect mapping performance to degrade the more semantically different the tweets are to the ``good'' examples and phrases provided by the experts. To achieve this, we look at evaluation metrics at different thresholds using the quartiles with respect to the proximity/similarity distribution. Results for $Q_1$ correspond to the 25\% most similar instances. For $Q_2$ to the 50\% most similar instances, and for $Q_3$ to the 75\% most similar instances. Note that these are continuous ranges and the quartiles serve as thresholds. 

To evaluate the impact of our neuro-symbolic mapping procedure (NeSym), we compared it against a nearest neighbors (NNs) approach that does not leverage conceptual frameworks and looks only at the language embedding of the tweets and theme examples and explanatory phrases. For the first iteration of Covid-19, we find that the approximate performance of the NeSym mapping at Q1 is better (+2 points) than the approximate full mapping for NNs, while increasing coverage x1.5. For immigration, we have an even more drastic result, having an approximate 15 point increase at a similar coverage gain. In both cases, experts were able to increase the number of themes in subsequent iterations\footnote{Due to effort required and cost, we only do a subsequent interactive session over the NeSym mapping.}. While the coverage increased in the second iteration for Covid, it decreased slightly for Immigration. For Covid, most of the coverage increase can be attributed to a single theme (\textit{Vax Efforts Progression}), which accounts for 20\% of the mapped data. In the case of Covid, this large jump in coverage is accompanied by a slight decrease in mapping performance. In the case of Immigration, we have the opposite effect: as the coverage decreases the performance improves, suggesting that the mapping gets stricter. This confirms the expected trade-off between coverage and quality. Depending on the needs of the final applications, experts could adjust their confidence thresholds. 

\begin{figure}[t]
    \centering
    \includegraphics[width=\columnwidth]{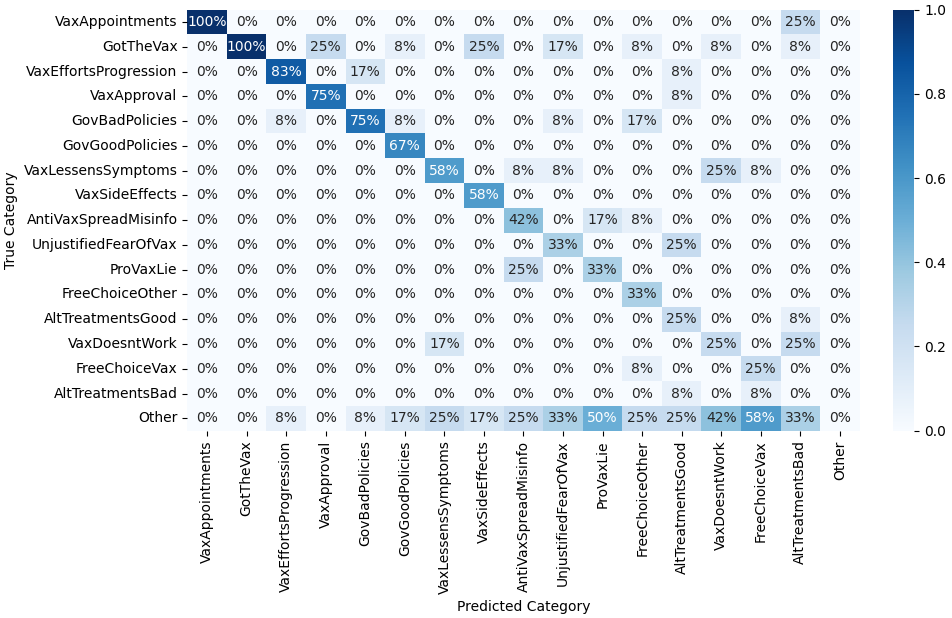}
    \caption{\textbf{Confusion matrix for Covid after second iteration}. Values are normalized over the predicted themes (cols), and sorted from best to worst.}
    \label{fig:cm_covid}
\end{figure}

To perform a fine-grained error analysis, we looked at the errors made by the model during manual validation. In Fig. \ref{fig:cm_covid} we show the confusion matrix for the Covid case. We find that the performance varies a lot, with some themes being more accurate than others. In some cases, we are good at capturing the general meaning of the theme but fail at grasping the stance similarities. For example, \textit{Anti Vax Spread Missinfo} gets confused with \textit{Pro Vax Lie}, where the difference is on who is doing the lying. In other cases, we find that themes that are close in meaning have some overlap (e.g. \textit{Alt Treatments} with \textit{Vax Doesn't Work}). We also find that unambiguous, neutral themes like \textit{Vax Appointments}, \textit{Got The Vax} and \textit{Vax Efforts Progression} have the highest performance. Lastly, we observe that for some errors, none of the existing themes are appropriate (Last row: \textit{Other}), suggesting that there are still undiscovered themes. Upon closer inspection, we found that the majority of these tweets are among the most distant from the theme embedding. The full distribution of \textit{Other} per interval can be observed in App. \ref{app:fine_grained_results}. We include the confusion matrix for immigration in App. \ref{app:fine_grained_results}.

Given our hypothesis that themes can be characterized by the strength of their relationship to high-level concepts, we consider mappings to be better if they are more cohesive. In the Covid case, we expect themes to have strong relationships to vaccination stance and morality frames. In the Immigration case, we expect themes to have strong relationships to the framing typologies. To measure this, we define a theme purity metric for each concept. For example, for stance this is defined as: $Purity_{stance} = \frac{1}{N} \sum_{t \in Themes} \max_{s \in Stance} |t \cap s|$. Namely, we take each theme cluster and count the number of data points from the most common stance value in said cluster (e.g. the number of data points that are \textit{anti-vax}). Then, we take the sum over all theme clusters and divide it by the number of data points. We do this for every concept, and average them to obtain the final averaged concept purity. In Tab.~\ref{tab:purity} we show the average concept purity for our mappings at each iteration in the interaction. We can see that the NeSym procedure results in higher purity with respect to the NNs procedure, even when significantly increasing coverage. This is unsurprising, as our method is designed to take advantage of the relationship between themes and concepts. Additionally, we include topic modeling baselines that do not involve any interaction, and find that interactive themes generally result in higher purity partitions than topics obtained using LDA. Details about the steps taken to obtain LDA topics can be found in App. \ref{app:topic_model_details}.

\begin{table}[t]
    \centering
    \resizebox{\columnwidth}{!}{%
    \begin{tabular}{ll|crl|cll}
    \toprule
        \multirow{2}{*}{Iter.} & Ground & \multicolumn{3}{c|}{Covid Vaccine} & \multicolumn{3}{c}{Immigration}  \\
        \cline{3-8}
        & Method & \# Thm & Cover & Purity & \# Thm & Cover & Purity  \\
        \midrule
        \rowcolor{LightGray} Baselines & LDA (Var. Bayes) & 9 & 39.8 & 63.72 & 13 & 26.8 & 57.14 \\
        \rowcolor{LightGray}  & LDA (Gibbs) &  & \textbf{79.8} & 63.90 &  & 55.9 & 54.86 \\
        \multirow{2}{*}{1} & NNs & & 9.3 & 68.81 & & 11.1 & 58.44 \\
        & NeSym & & 54.3 & \textbf{69.97} & & \textbf{65.8} & \textbf{61.72} \\
        \midrule
        \rowcolor{LightGray} Baselines & LDA (Var. Bayes) & 16 & 26.1 & 65.02 & 19 & 18.3 & 57.94 \\
        \rowcolor{LightGray}  & LDA (Gibbs) &  & 73.1 & 65.14 &  & 46.8 & \textbf{59.25} \\
        2 & NeSym & & \textbf{84.3} & \textbf{65.50}  & & \textbf{59.6} & 59.19 \\
        \bottomrule
    \end{tabular}}
    \caption{\textbf{Dataset Coverage and Avg. Concept Purity}. For LDA, we assigned a tweet to its most probable topic if the probability was $\geq 0.5$.}
    \label{tab:purity}
\end{table}

\begin{figure*}[ht]
    \centering
    \includegraphics[width=\textwidth]{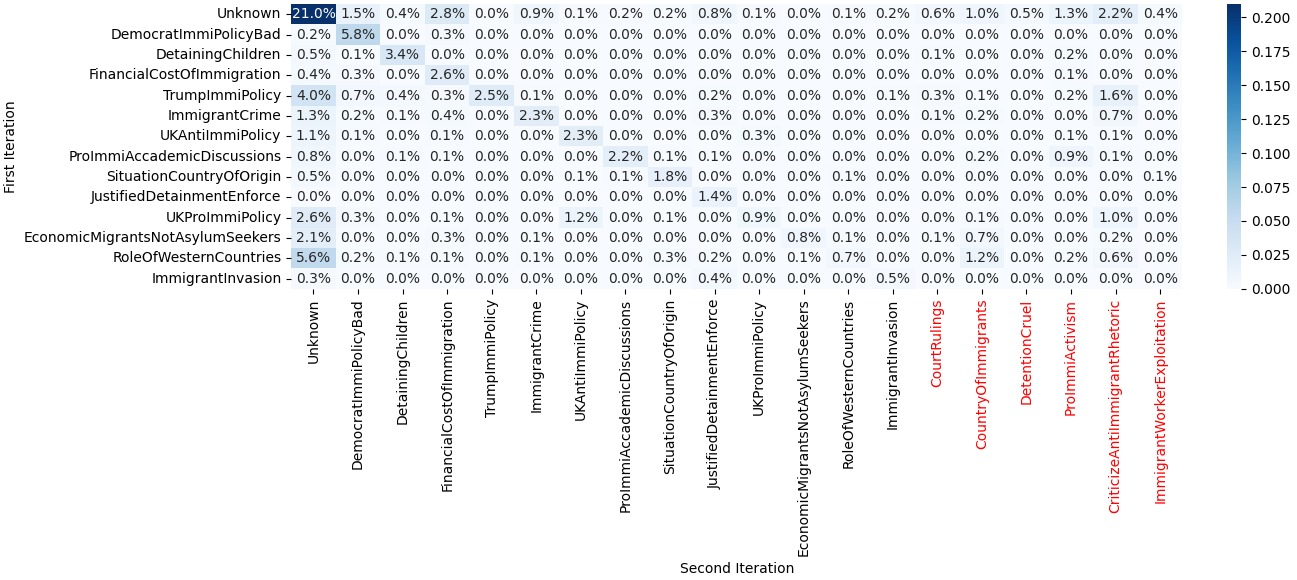}
    \caption{\textbf{Shifting predictions for Immigration}. Themes added during second iteration are shown in red, and values are normalized over the full population.}
    \label{fig:shifting_preds_immi}
\end{figure*}

\paragraph{Effects of Consecutive Iterations} In Fig. \ref{fig:general} we observed different behaviors in subsequent iterations with respect to coverage and performance. To further inspect this phenomenon, we looked at the tweets that shifted predictions between the first and second iterations. Fig. \ref{fig:shifting_preds_immi} shows this analysis for Immigration. Here, we find that a considerable number of tweets that were assigned to a theme in the first iteration, were unmatched (i.e. moved to the \textit{Unknown}) in the second iteration. This behavior explains the decrease in coverage. Upon closer inspection, we found that the majority of these unmatched tweets corresponded to assignments that were in the last and second to last intervals with respect to their similarity to the theme embedding. We also observed a non-trivial movement from the \textit{Unknown} to the new themes (shown in red), as well as some shifts between old themes and new themes that seem reasonable. For example, 1.2\% of the total tweets moved from \textit{Role of Western Countries} to \textit{Country of Immigrants}, 1\% moved from \textit{Academic Discussions} to \textit{Activism}, and close to 3\% of tweets moved from \textit{Trump Policy} and \textit{UK Policy} to \textit{Criticize Anti Immigrant Rhetoric}. This behavior, coupled with the increase in performance observed, suggests that as new themes are added, tweets move to a closer fit. In App. \ref{app:shifting_preds} we include the shift matrix for Covid, as well as the distribution of the unmatched tweets with respect to their semantic similarity to the theme embedding. For Covid, we observe that the increase in coverage is mostly attributed to the addition of the \textit{Vax Efforts Progression} theme, which encompasses all mentions to vaccine development and roll-out. Otherwise, a similar shifting behavior can be appreciated. 

\begin{figure}[t]
%
  \centering
 \includegraphics[width=\columnwidth]{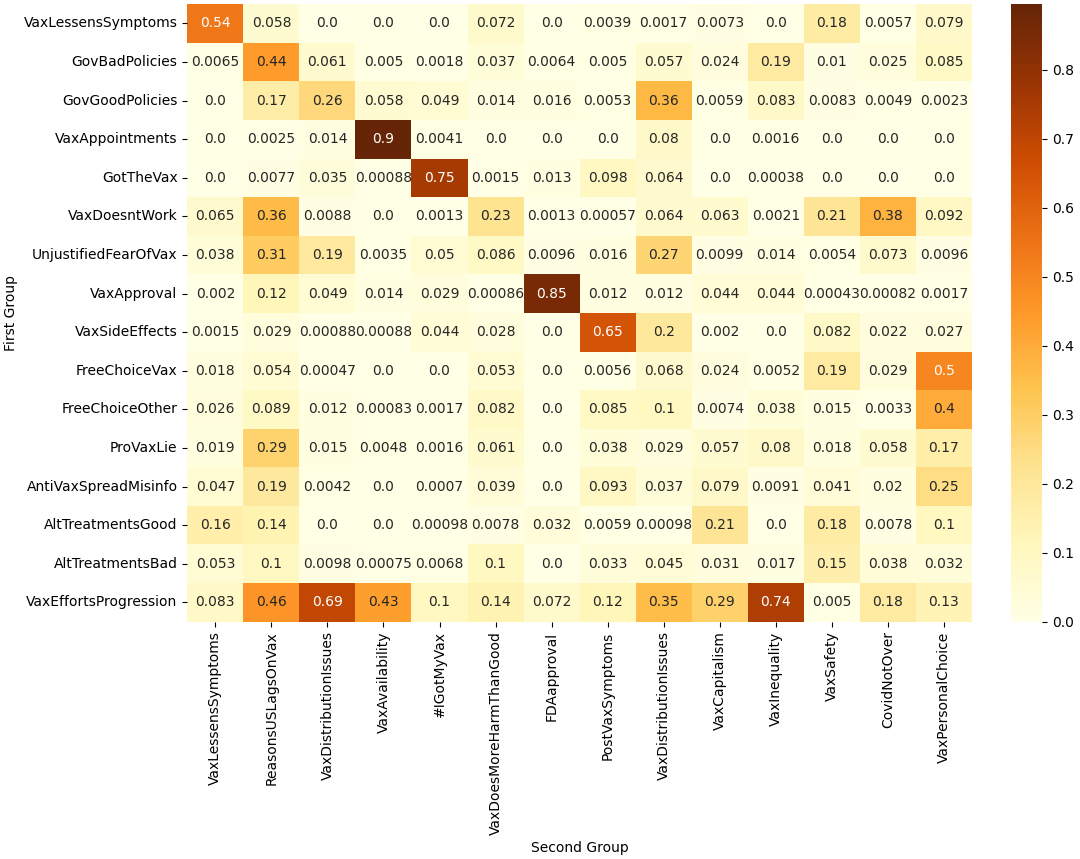}
\caption{Theme Overlap Coefficient Heatmap between Different Groups of Experts}\label{fig:group_overlap}
\end{figure}

\paragraph{Consistency between Different Expert Groups} To study the subjectivity of experts and its impact on the resulting themes, we performed two parallel studies on the Covid corpus. For each study, a different group of experts performed two rounds of interaction following the protocol outlined on Sec. \ref{sec:framework}. The side-by-side comparisons of the two studies can be observed in Tab.~\ref{tab:different_groups}. We find that the second group of experts is able to obtain higher coverage and higher concept purity with a slightly reduced number of themes. To further inspect this phenomenon, as well as the similarities and differences between the two sets of themes, we plot the overlap coefficients between the theme-to-tweet mappings in Fig.~\ref{fig:group_overlap}. We use the Szymkiewicz–Simpson coefficient, which measures the overlap between two finite sets and is defined as: $overlap(X,Y) = \frac{|X \cap Y|}{min(|X|,|Y|)}$.  

\begin{table}[t]\scriptsize
    \centering
    \resizebox{0.8\columnwidth}{!}{%
    \begin{tabular}{clcc}
    \toprule
    Iter.   & Metric & Group 1 & Group 2 \\
    \midrule
    1     &  Num Themes & 9 & 8 \\
          &  Coverage & 54.30 & 61.80 \\
          &  Stance Purity & 83.18 & 87.43 \\
          &  Moral Frame Purity & 56.75 & 65.52 \\
    \midrule
    2     &  Num Themes & 16 & 14 \\
          &  Coverage & 84.30 & 85.90 \\
          &  Stance Purity & 80.12 & 84.31  \\
          &  Moral Frame Purity & 50.88 & 52.17 \\
    \bottomrule
    \end{tabular}}
    \caption{Two Different Groups of Experts on  \textbf{Covid}}
    \label{tab:different_groups}
\end{table}

In cases where we observe high overlap between the two groups, we find that there is essentially a word-for-word match between the two discovered themes. For example, \textit{Vax Lessens Symptoms}, which was surprisingly named the same by the two groups, as well as \textit{Vax Availability} vs. \textit{Vax Appointments}, \textit{Got The Vax} vs. \textit{I Got My Vax}, and \textit{Vax Side Effects} vs. \textit{Post Vax Symptoms}. In other cases, we find that different groups came up with themes that have some conceptual (and literal) overlap, but that span different sub-segments of the data. For example, we see that the theme \textit{Reasons the US Lags On Vax} defined by the second group, has overlap with different related themes in the first group, such as: \textit{Gov. Bad Policies}, \textit{Vax Efforts Progression}, and \textit{Unjustified Fear of Vax}. Similarly, while the second group defined a single theme \textit{Vax Personal Choice}, the first group attempted to break down references to personal choices between those direclty related to taking the vaccine (\textit{Free Choice Vax}), and those that use the vaccine as analogies for other topics, like abortion (\textit{Free Choice Other}). While some themes are clearly present in the data and identified by the two groups, we see that subjective decisions can influence the results. The first group was inclined to finer grained themes (with the exception of \textit{Vax Efforts Progression}), while the second group seemed to prefer more general themes. In future work, we would like to study how the variations observed with our approach compare to the variations encountered when experts follow fully manual procedures, as well the impact of the crowd vs. experts working alone. 

\begin{figure}[ht]
    \centering
    \includegraphics[width=\columnwidth]{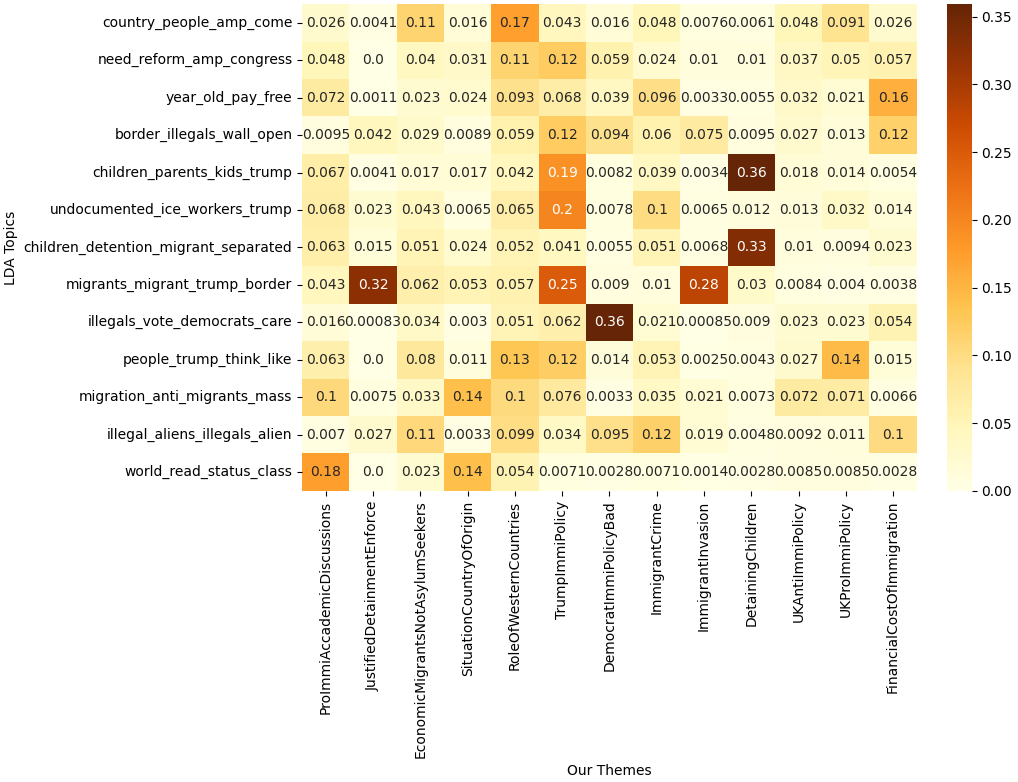}
    \caption{Overlap Coefficients between LDA Var. Bayes and our Themes (First Iter. \textbf{Immigration}). 
    }
    \label{fig:lda_theme_overlap_iter1_immi}
\end{figure}

\paragraph{Abstract Themes vs. Word-level Topics} To get more insight into the differences between topics based on word distributions and our themes, we looked at the overlap coefficients between topics obtained using LDA and our themes. Fig. \ref{fig:lda_theme_overlap_iter1_immi} shows the coefficients for Immigration. While some overlap exists, the coefficients are never too high (a max. of 0.35). One interesting finding is that most of our themes span multiple related topics. For example, we find that \textit{Trump Policy} has similar overlap with \textit{undocumented\_ice\_workers\_trump}, \textit{migrants\_migrant\_trump\_border}, and \textit{children\_parent\_kids\_trump}. While all of these topics discuss Trump policies, they make reference to different aspects:  workers, the border and families. This supports our hypothesis that our themes are more abstract in nature, and that they capture conceptual similarities beyond word distributions. Overlap coefficients for Gibbs sampling, Covid, and subsequent iterations can be seen in App.~\ref{app:lda_themes}.

\section{Limitations}

The study presented in this paper has three main limitations. (1) While the design of the framework does not prohibit the utilization of longer textual forms, the two case studies presented deal with short texts. When dealing with longer text forms, we need to consider the cognitive load of having experts look at groups of instances. In our ongoing work, we employ strategies such as summarization, highlighting and other visualization techniques to deal with these challenges. (2) In the studies presented, qualitative researchers worked in groups to identify themes. Our goal in comparing two independent groups of researchers was to evaluate the degree of subjectivity by observing if the themes identified by the two groups would diverge. This setup might not always be realistic, as a lot of times qualitative researchers work independently or asynchronously. In the future, we will explore the effect of the crowd in minimizing subjectivity, as well as the role that the computational tools play in more challenging settings. (3) Finally, we did not include a comprehensive user study to gather input from the experts about their experience with our framework. We consider this to be an important next step, and we are actively working in this direction. 
\section{Summary}

We presented a concept-driven framework for uncovering latent themes in text collections. Our framework expands the definitions of a theme to account for theoretically informed concepts that generalize beyond word co-occurrence patterns. We suggest an interactive protocol that allows domain experts to interact with the data and provide feedback at different levels of abstraction. We performed an exhaustive evaluation using two case studies and different groups of experts. Additionally, we contrasted the extracted themes against the output of traditional topic models, and showed that they are better at capturing conceptual similarities that go beyond word distributions.

\bibliography{anthology,custom}
\bibliographystyle{acl_natbib}

\newpage

\appendix

\section{Appendix}
\label{sec:appendix}


\subsection{Tool Screenshots}\label{app:tool}

\subsubsection{Exploratory Operations}\label{app:exploratory}

\begin{figure}[H]
    \centering
    \includegraphics[width=0.5\columnwidth]{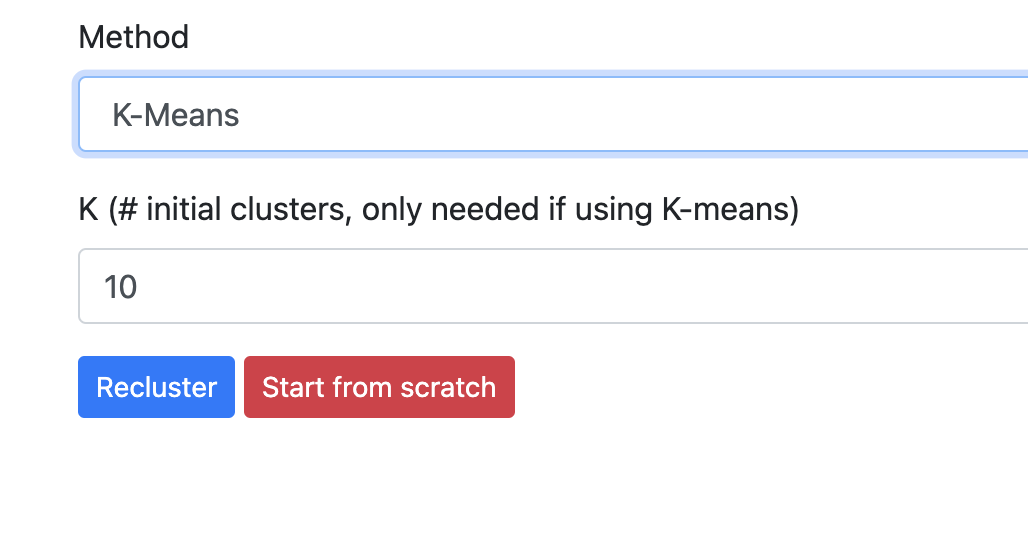}%
    \caption{Cluster Instances}
    \label{fig:cluster_tweets}
\end{figure}

\begin{figure}[H]
    \centering
    \includegraphics[width=\columnwidth]{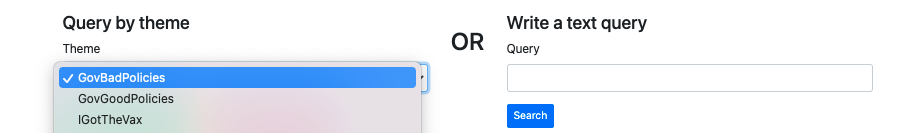}%
    \caption{Text-based Queries}
    \label{fig:query_tweets}
\end{figure}

\begin{figure}[H]
    \centering
    \includegraphics[width=\columnwidth]{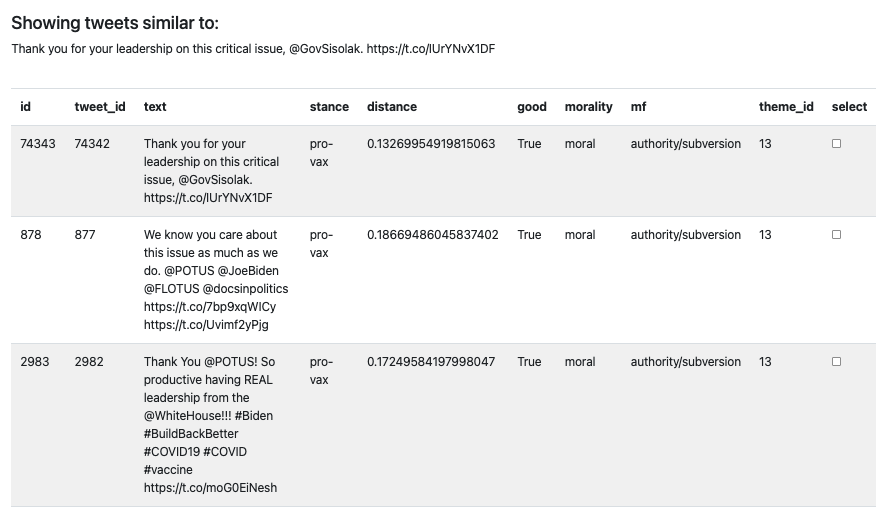}%
    \caption{Finding Similar Tweets}
    \label{fig:explore_similar}
\end{figure}

\begin{figure}[H]
    \centering
    \includegraphics[width=\columnwidth]{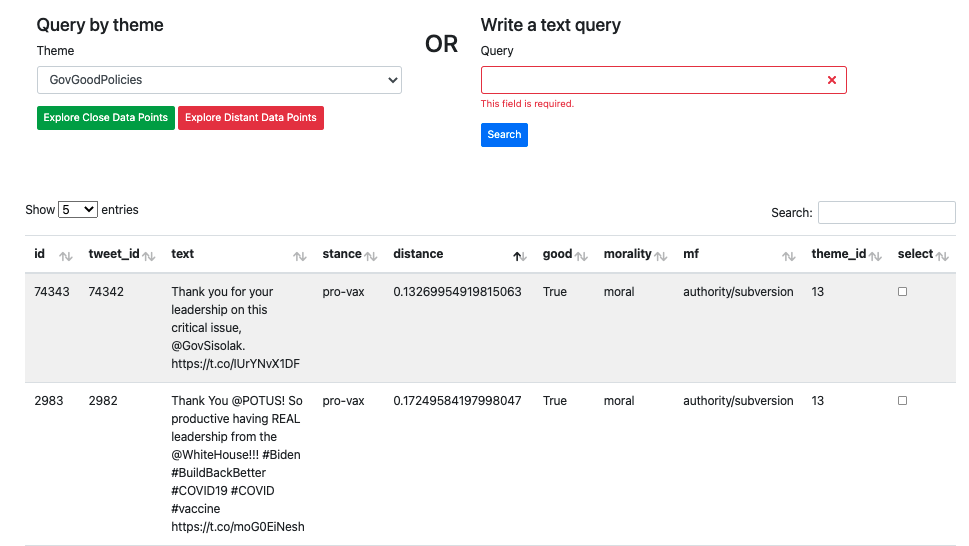}%
    \caption{Listing Arguments and Examples}
    \label{fig:show_instances}
\end{figure}

\begin{figure}[H]
    \centering
    \includegraphics[width=0.5\columnwidth]{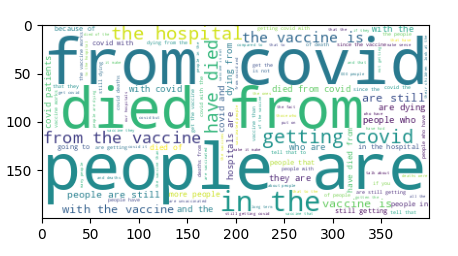}%
    \caption{Visualizing Local Explanations: Word Cloud Example for \textit{The Vaccine Doesn't Work}}
    \label{fig:wordcloud_example}
\end{figure}

\begin{figure*}[h]
\centering
\begin{subfigure}{0.5\linewidth}
  \centering
  \includegraphics[width=\linewidth]{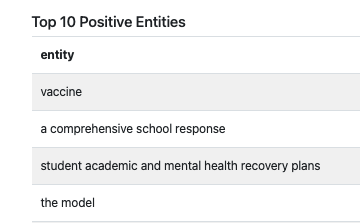}
  \caption{Top Positive Entities}
\end{subfigure}
\begin{subfigure}{0.4\linewidth}
  \centering
  \includegraphics[width=\linewidth]{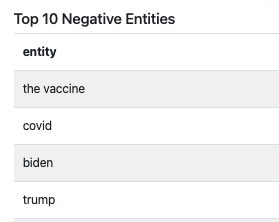}
  \caption{Top Negative Entities}
\end{subfigure}

\caption{Visualizing Local Explanations: Most Frequent Positive and Negative Entities for \textit{Bad Governmental Policies}}
\label{fig:top_entities}
\end{figure*}

\begin{figure}[H]
\centering
\begin{subfigure}{0.5\columnwidth}
  \centering
  \includegraphics[width=0.9\textwidth]{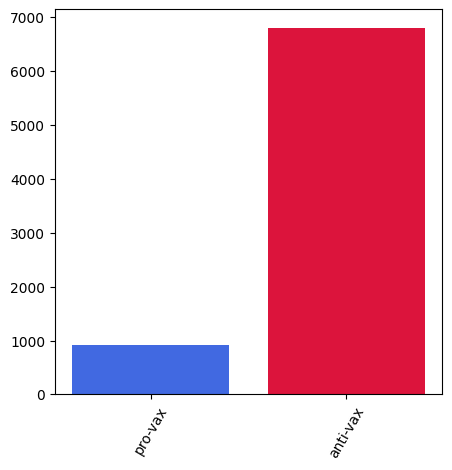}
  \caption{Stance}
\end{subfigure}%
\begin{subfigure}{0.5\columnwidth}
  \centering
  \includegraphics[width=0.9\textwidth]{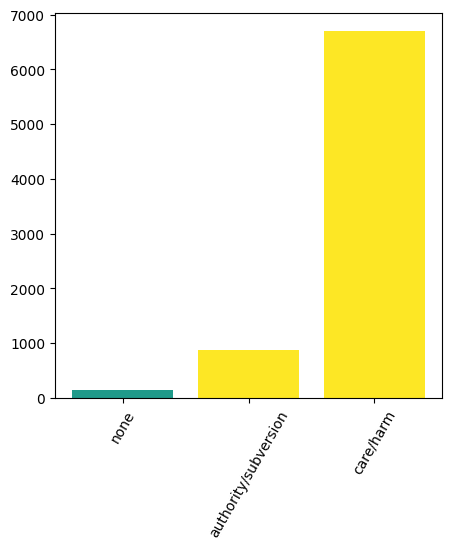}
  \caption{Moral Foundation}
\end{subfigure}

\caption{Visualizing Local Explanations: Attribute Distribution for \textit{The Vaccine Doesn't Work}}
\end{figure}

\begin{figure}[H]
    \centering
    \includegraphics[width=\columnwidth]{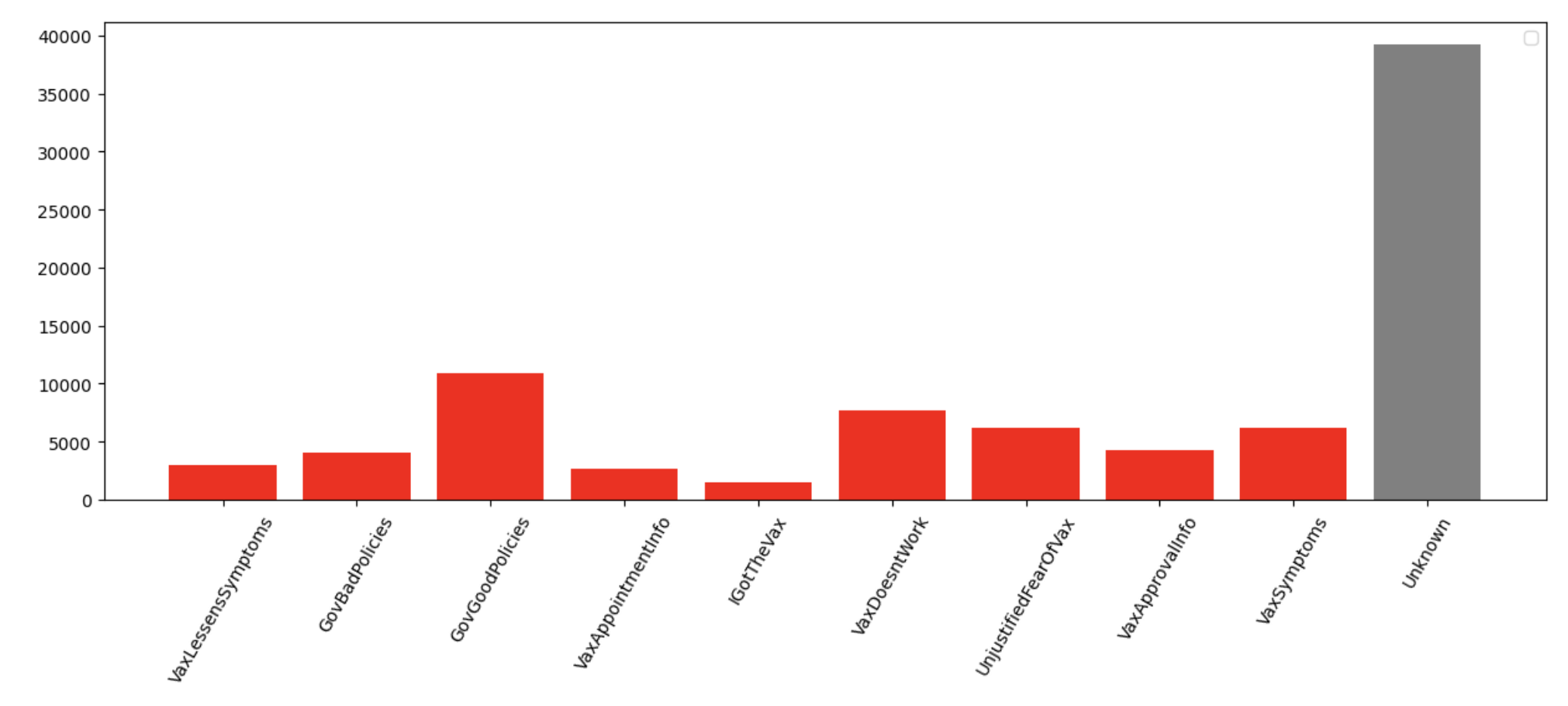}%
    \caption{Visualizing Global Explanations: Theme Distribution}
    \label{fig:theme_distribution}
\end{figure}

\begin{figure}[H]
\centering
  \includegraphics[width=0.3\textwidth]{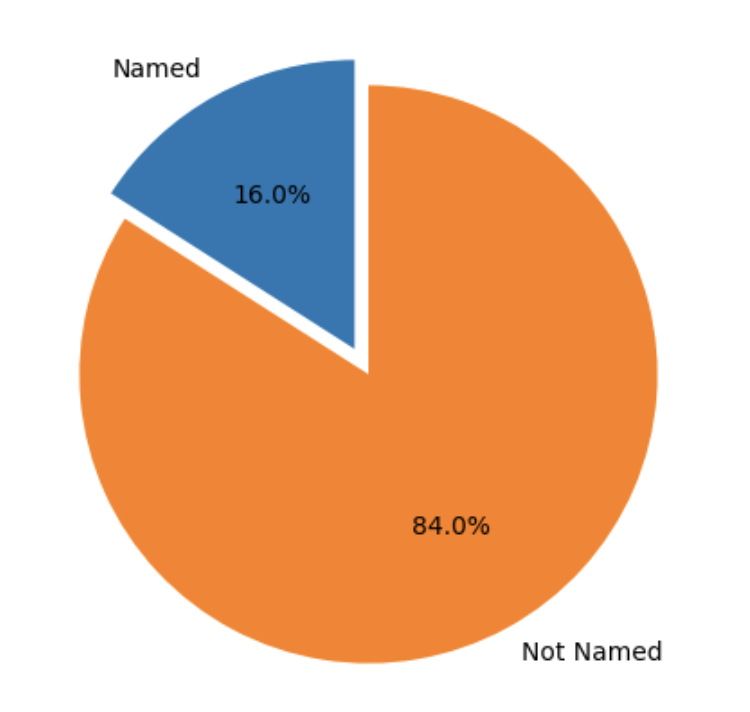}
  \caption{Visualizing Global Explanations: Coverage}\label{fig:coverage_example}
\end{figure}

\begin{figure}[H]
\centering
  \includegraphics[width=\columnwidth]{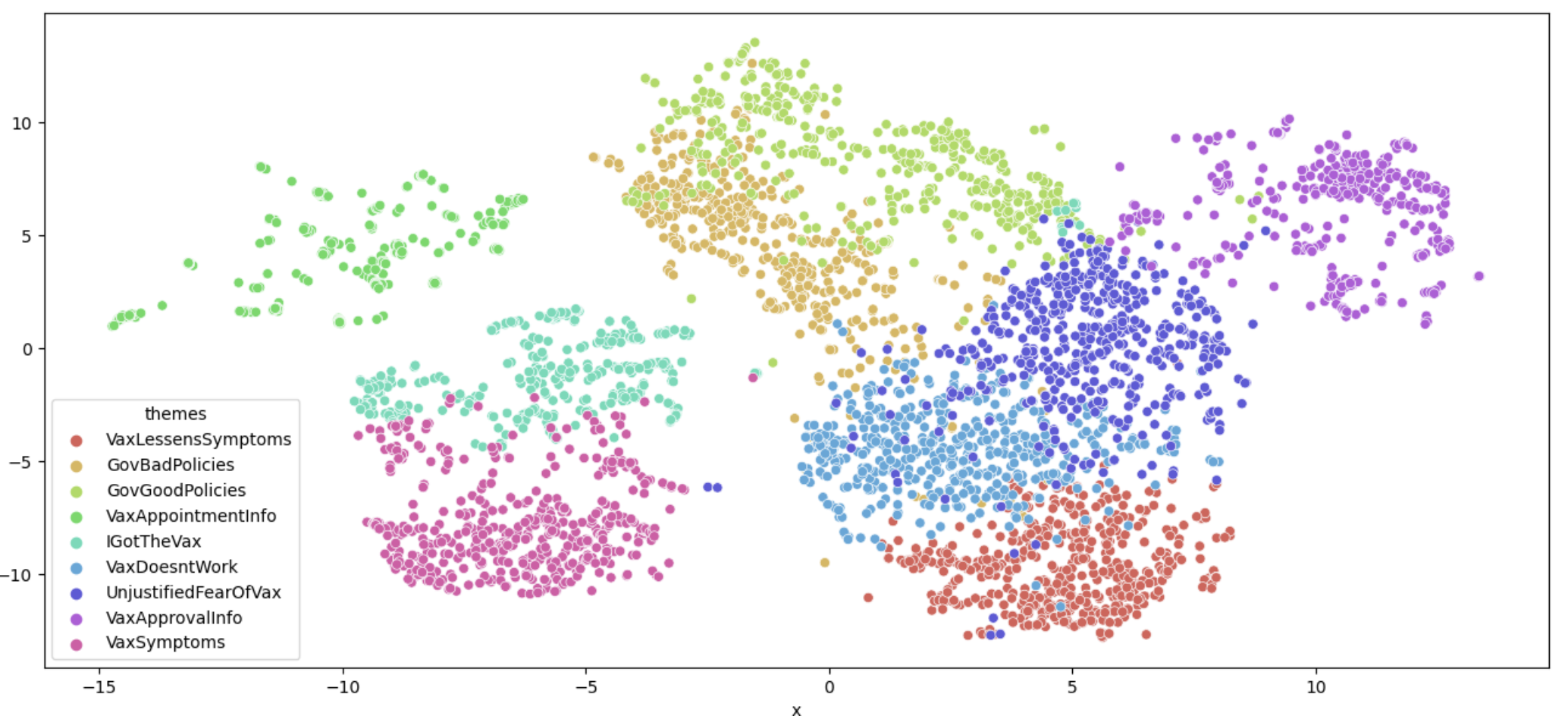}
  \caption{Visualizing Global Explanations: 2D t-SNE}\label{fig:tsne}
\end{figure}

\subsubsection{Intervention Operations}\label{app:intervention}

\begin{figure}[H]
\centering
  \includegraphics[width=\columnwidth]{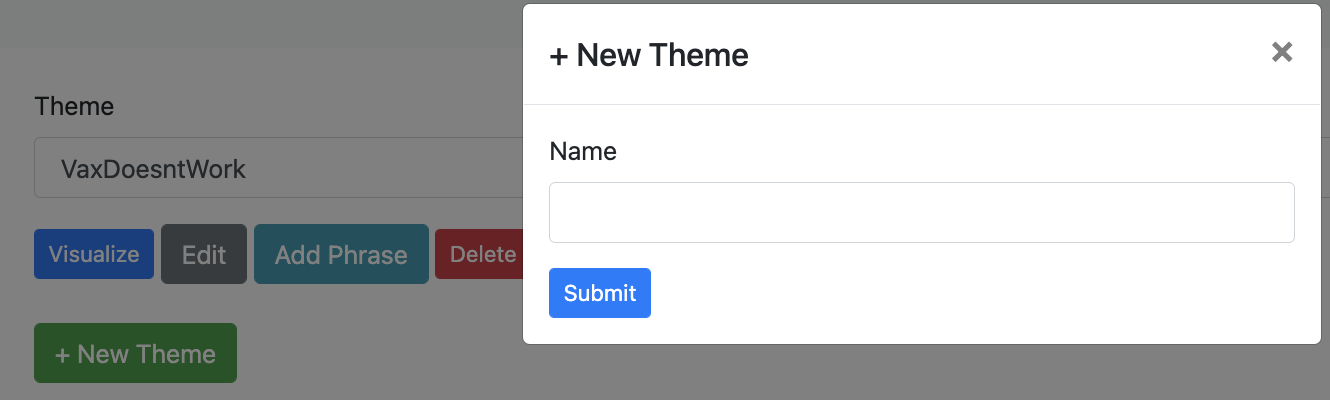}
  \caption{Adding New Themes}\label{fig:new_theme}
\end{figure}

\begin{figure}[H]
\centering
  \includegraphics[width=\columnwidth]{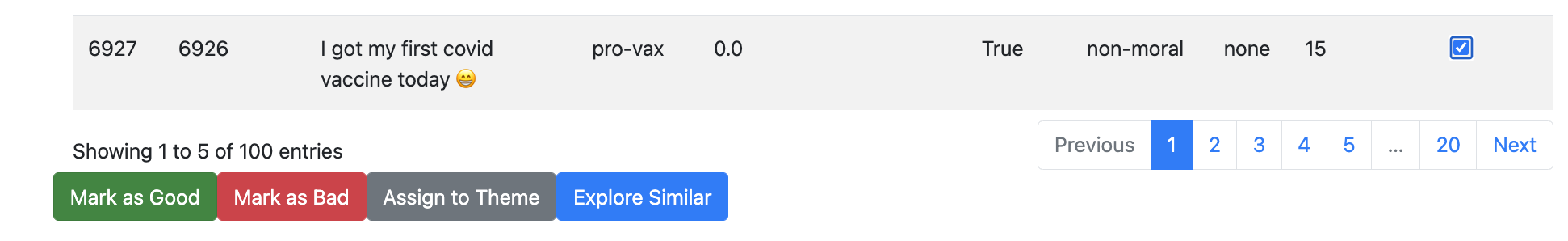}
  \caption{Marking Instances as \textit{Good}}\label{fig:mark_good_bad}
\end{figure}

\begin{figure}[H]
\centering
  \includegraphics[width=\columnwidth]{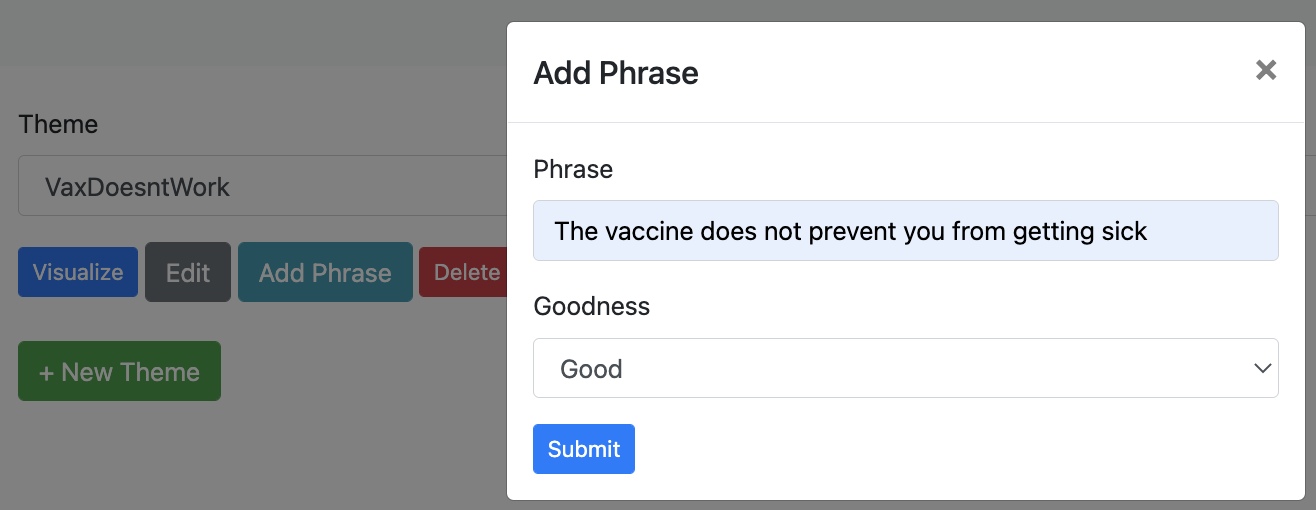}
  \caption{Adding \textit{Good} Examples}\label{fig:add_phrase}
\end{figure}

\begin{figure}[ht]
\centering
  \includegraphics[width=\columnwidth]{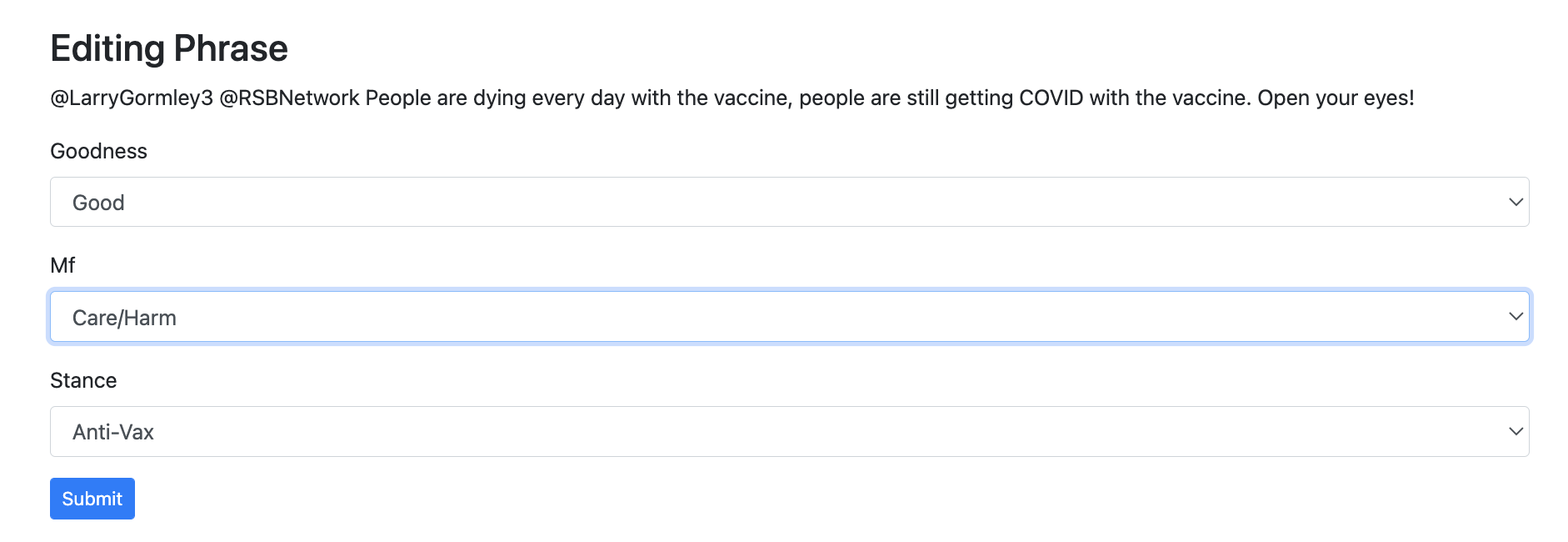}
  \caption{Correcting Attributes - Stances and Moral Foundations}\label{fig:edit_judgements}
\end{figure}

\subsection{Interactive Sessions for Covid: First Group of Experts}\label{app:covid_sessions_first}

Table \ref{tab:session_covid_1} and \ref{tab:session_covid_2} outline the patterns discovered by the the first group of experts on the first a second iteration, respectively.

\begin{table*}
\centering
    \resizebox{\textwidth}{!}{%
    \begin{tabular}{|l|l|l|}
      \hline
      \thead{\textbf{Cluster}} & \thead{\textbf{Experts Rationale}} & \thead{\textbf{New Named Themes}} \\
      \hline
        K-Means 0 & Discusses what the vaccine can and cannot do. & VaxLessensSymptoms \\
        & Emphasis in reducing COVID-19 symptoms in case of infection &  \\
        & (``like a bad cold''). Contains tweets with both stances. & \\
        \hline
        K-Means 1 & A lot of mentions to political entities. & GovBadPolicies \\
        & Politicians get in the way of public safety & \\
        \hline
        K-Means 2 & A lot of tweets with mentions and links. & GovGoodPolicies \\
        & Not a lot of textual context. & \\
        & Some examples thanking and praising governmental policies. & \\
        & \textbf{Theme added upon inspecting similar tweets} & \\
        \hline
        K-Means 3 & Overarching theme related to vaccine rollout. & \\  
        & Mentions to pharmacies that can distribute, & - \\
        & distribution in certain states, & \\
        & places with unfulfilled vax appointments. & \\
        & \textbf{Too broad to create a theme} & \\
        \hline
        K-Means 4 & Broadcast of vaccine appointments. & VaxAppointments \\
        & Which places you can get vaccine appointments at. & \\
        \hline
        K-Means 5 & “I got my vaccine” type tweets & GotTheVax \\
        \hline
        K-Means 6 & Mixed cluster, not a clear theme in centroid. & VaxDoesntWork \\
        & Two prominent flavors: the vaccine not working and & UnjustifiedFearOfVax \\
        & people complaining about those who are scared of vaccine. & \\
        \hline
        K-Means 7 & Tweets look the same as K-Means 5 & - \\
        \hline
        K-Means 8 & Tweets about development and approval of vaccines & VaxApproval \\
        \hline
        K-Means 9 & Tweets related to common vaccine side-effects & VaxSideEffects \\
        \hline
    \end{tabular}}
        \caption{\textbf{First Iteration}: Patterns Identified in Initial Clusters and Resulting Themes}
    \label{tab:session_covid_1}
\end{table*}

\begin{table*}
\centering
    \resizebox{\textwidth}{!}{%
    \begin{tabular}{|l|l|l|}
      \hline
      \thead{\textbf{Cluster}} & \thead{\textbf{Experts Rationale}} & \thead{\textbf{New Named Themes}} \\
      \hline
        K-Means 0 & Tweets weighting health benefits/risks, but different arguments. & \\
        & (e.g. it works, doesn't work, makes things worse...) & - \\
        & \textbf{Too broad to create a theme}. &  \\
        \hline
        K-Means 1 & Messy cluster, relies on link for information. & - \\
        \hline
        K-Means 2 & Relies on link for information. & - \\
        \hline
        K-Means 3 & A lot of mentions to government lying and misinformation. & AntiVaxSpreadMisinfo \\
        & ``misinformation'' is used when blaming antivax people. & ProVaxLie \\
        & ``experts and government are lying'' is used on the other side. & AltTreatmentsGood \\
        & References to alt-treatments on both sides. & AltTreatmentsBad \\
        & \textbf{Text lookup ``give “us the real meds'', ``covid meds'' } &  \\
        \hline
        K-Means 4 & Some examples are a good fit for old theme, VaxDoesntWork. & - \\
        & \textbf{Other than that no coherent theme.}&  \\
        \hline
        K-Means 5 & Tweets about free will and choice. & FreeChoiceVax \\
        & \textbf{Text lookup ``big gov'', ``free choice'', ``my body my choice''} & FreeChoiceOther \\
        & Case ``my body my choice'' - a lot of mentions to abortion & \\
        & People using covid as a metaphor for other issues. & \\ 
        \hline
        K-Means 6 & Almost exclusively mentions to stories and news. 
& - \\
        \hline
        K-Means 7 & Availability of the vaccine, policy. & VaxEffortsProgression \\
        & Not judgement of good or bad, but of how well it progresses. &  \\
        \hline
        K-Means 8 & Assign to previous theme GotTheVax  & - \\
        \hline
        K-Means 9 & Vaccine side effects.  & - \\
        & Assign to previous theme, VaxSymptoms & \\
        \hline
    \end{tabular}}
        \caption{\textbf{Second Iteration}: Patterns Identified in Subsequent Clusters and Resulting Themes}
    \label{tab:session_covid_2}
\end{table*}

\subsection{Interactive Sessions for Covid: Second Group of Experts}\label{app:covid_sessions_second}

Table \ref{tab:first_session_second_group} and \ref{tab:second_session_second_group} outline the patterns discovered by the second group of experts on the first a second iteration, respectively. 

\begin{table*}
\centering
    \resizebox{\textwidth}{!}{%
    \begin{tabular}{|l|l|l|}
      \hline
      \thead{\textbf{Cluster}} & \thead{\textbf{Experts Rationale}} & \thead{\textbf{New Named Themes}} \\
      \hline
        K-Means 0 & People asking people to get vaccinated. & VaxLessensSymptoms \\
        & Some skeptical but acknowledge it reduces symptoms. &  \\
        & It works but it has limitations. & \\
        & More specifically, it lessens the symptoms. & \\
        \hline
        K-Means 1 & Republicans have hurt the vax rate in the US.  & ReasonsUSLagsOnVax \\
        &  Finding someone (or some party) to blame.  & \\
        & Politicians are hurting people with policy. & \\
        & Vaccine in the US is behind, trying to explain why & \\
        \hline
        K-Means 2 & A lot of them are just replies. & - \\
        & Cluster is for links and usernames. &
        \\
        \hline
        K-Means 3 & Availability and distribution of the vaccine. & VaxDistributionIssuesDueToLocalPolicy \\ 
        & How stances of people in different states affect it.  & \\  
        & Vaccine distribution issues due to local policy. & \\
        \hline
        K-Means 4 & Clear cluster. Vaccine info, availability info.  & VaxAvailabilityInfo \\
        \hline
        K-Means 5 & Testimonials, \#IGotMyVax & \#IGotMyVax \\
        \hline
        K-Means 6 & Some themes match the vaccine lessens symptoms. & VaxDoesMoreHarmThanGood \\
        & Other theme: no need to get the vaccine, it doesn’t work. & \\
        & Vaccine does more harm than good. & \\
        \hline
        K-Means 7 & Same as K-means 5 & - \\
        \hline
        K-Means 8 & About covid vaccine updates. FDA approval. & FDAApproval \\
        & In other cases it depends on the content on the link. & \\
        & So you can’t really tell. & \\
        \hline
        K-Means 9 & Obvious. Vaccine symptoms, vaccine effects. & PostVaxSymptoms \\
        & Post vaccination symptoms.  &  \\
        \hline
    \end{tabular}}
        \caption{\textbf{Second Group's First Iteration}: Patterns Identified in Initial Clusters and Resulting Themes}
    \label{tab:first_session_second_group}
\end{table*}

\begin{table*}
\centering
    \resizebox{\textwidth}{!}{%
    \begin{tabular}{|l|l|l|}
      \hline
      \thead{\textbf{Cluster}} & \thead{\textbf{Experts Rationale}} & \thead{\textbf{New Named Themes}} \\
      \hline
      K-Means 0 & Links and promotions & - \\ \hline
      K-Means 1 & Looks like previous theme IGotMyVax, assign them. & -\\ \hline
      K-Means 2 & Very short tweets with links, and no context.  & - \\
      & Could be availability but not sure. Decided against adding theme & \\
      \hline
      K-Means 3 & Two themes observed. One old one, regarding VaxAvailabilityInfo. & VaxDistributionIssues \\  
      & One new one, getting vaccines is difficult. Not related to local policy. &  \\
      & \textbf{Decided against merging with previous theme} & \\
      \hline
      K-Means 4 & A lot of talk about skepticism regarding the vaccine.  & VaxCapitalism \\
      & Some good matches to previous MoreHarmThanGood, assign them. & VaxInequality  \\
      & Mentions to profiting from the vaccine. &  \\
      & \textbf{Look for similar instances to mentions of profits} & \\
      & \textbf{Text look up for "vaccine getting rich"} & \\
      & Mentions to redlining, implications of inequality & \\
      & \textbf{Text look up for "vaccine inequality"} & \\
      & Lots of mentions to racial and monetary inequalities in access to vaccine & \\
      \hline
      K-Means 5 & Both PostVaxSymptoms and IGotMyVax examples, assign them. & - \\  \hline
      K-Means 6 & Mentions to vaccine safety. Weighting the safety/risks of the vaccine & VaxSafety \\  \hline
      K-Means 7 & A lot of discussion about the pandemic not being over & CovidNotOver \\  
      & Discussion on whether to open back up or not & \\
      \hline
      K-Means 8 & Repetitions, IGotMyVax. Assign them. & - \\  \hline
      K-Means 9 & Mentions to mandates. & VaxPersonalChoice \\
      & The vaccine should be a personal choice, mandates should not be there. & \\
      & Different reasons: personal choice, no proof of whether it works. & \\
      & For no proof, assign to previous MoreHarmThanGood & \\
      \hline
 \end{tabular}}
        \caption{\textbf{Second Group's Second Iteration}: Patterns Identified in Subsequent Clusters and Resulting Themes}

    \label{tab:second_session_second_group}
\end{table*}

\subsection{Interactive Sessions for Immigration}\label{app:immi_sessions}

Table \ref{tab:first_session_immi} and \ref{tab:second_session_immi} outline the patterns discovered by the experts for immigration.

\begin{table*}[ht]
\centering
    \resizebox{\textwidth}{!}{%
    \begin{tabular}{|l|l|l|}
      \hline
      \thead{\textbf{Cluster}} & \thead{\textbf{Experts Rationale}} & \thead{\textbf{New Named Themes}} \\
      \hline
      K-Means 0 & Headlines, coverage. Some have an agenda (pro) & AcademicDiscussions \\ 
      & Others are very academic and research-oriented & \\
      & Opinion pieces. & \\
      \hline
      K-Means 1 & Talking about apprehending immigrants at the border & JustifiedDetainmentEnforce \\
      & Some report about the border but no stance. Deportation. & \\
      & Leaning negative towards immigrants. & \\
      \hline
      K-Means 2 & Less US-centric, more general. & EconomicMigrantsNotAsylumSeekers \\
      & Talking about immigration as a global issue & SituationCountryOfOrigin \\
      & Humanitarian issues, mentions to refugees, forced migration & RoleOfWesternCountries \\
      & Situation in country of origin that motivates immigration & \\
      & Mentions to how the west is responsible & \\
      & The role of the target countries in destabilizing countries  & \\
      & Mentions to economic migrants. & \\ 
      & \textbf{Look up for "economic work migrants", "asylum seekers"} & \\
      \hline
      K-Means 3 & About Trump. Trump immigration policy. & TrumpImmiPolicy \\
      & Politicizing immigration. & \\
      \hline
      K-Means 4 & Attacking democrats. & DemocratImmiPolicyBad \\
      & A lot of mentions to democrats wanting votes & \\
      & Common threads is democrats are bad & \\
      \hline
      K-Means 5 & Lacks context, lots of usernames. & ImmigrantInvasion \\  
      & Not a cohesive theme. Both pro and con, and vague.  & ImmigrantCrime \\
      & Some mentions to invasion. \textbf{Look for "illegal immigrants invade"} & \\
      & Mentions to caravan, massive exodus of people. Mentions to crime. & \\
      &\textbf{Look for immigrants murder, immigrants dangerous.} & \\
      & A lot of tweets linking immigrants to crime & \\
      \hline
      K-Means 6 & Looks very varied. Not cohesive. & - \\  \hline
      K-Means 7 & Very cohesive. Mentions to detaining children, families. & DetainingChildren \\  \hline
      K-Means 8 & All tweets are about the UK and Britain. & UKProImmiPolicy\\  
      & Both pro and anti immigration. & UKAntiImmiPolicy  \\ 
      & Only common theme is the UK. Almost exclusively policy/politics & \\
      \hline
      K-Means 9 & Economic cost of immigration. & FinacialCostOfImmigration \\ 
      & Immigration is bad for the US economy & \\
      & Some about crime, and democrats. Assign to existing themes. & \\
      \hline
 \end{tabular}}
        \caption{\textbf{First Iteration Immigration}: Patterns Identified in Initial Clusters and Resulting Themes}

    \label{tab:first_session_immi}
\end{table*}

\begin{table*}[ht]
\centering
    \resizebox{\textwidth}{!}{%
    \begin{tabular}{|l|l|l|}
      \hline
      \thead{\textbf{Cluster}} & \thead{\textbf{Experts Rationale}} & \thead{\textbf{New Named Themes}} \\
      \hline
      K-Means 0 & Legal decisions and rulings. & CourtRulings \\ 
      & Both pro and anti immigration rulings & \\
      & Not a single event, but cohesively talking about rulings & \\
      \hline
      K-Means 1 & The same tweet reworded and tweeted at different people & ImmigrantWorkerExploitation \\
      & Talks about worker exploitation, and Cesar Chavez. & \\
      & \textbf{Look up for "exploitation"}. Mentions to workers and wages & \\
      & \textbf{Look up for "cheap labor"} & \\
      \hline
      K-Means 2 & Blaming Trump for being irresponsible & CriticizeAntiImmigrantRhetoric \\
      & Criticizing his rhetoric. Mentions to hateful speech & \\ 
      & About the rhetoric rather than policy. Mentions to racist language & \\
      & Others about policy, added to previous TrumpImmiPolicy theme & \\
      \hline
      K-Means 3 & Nation of immigrants. Identity, we are all immigrants & CountryOfImmigrants \\
      \hline
      K-Means 4 & Organizing. Call to action. Skews pro. language of rights and liberties. & ProImmiActivism \\
      & We are here, we demand, sign here. \textbf{Look up "ACLU", "rights for immigrants"} & \\
      \hline
      K-Means 5 & A lot of mentions to numbers and stats. Short URLs. Headlines. & - \\  \hline
      K-Means 6 & A lot of usernames. Bad policies, criticizing policies on both sides. & - \\  
      & Send them to either DemocratImmiPolicyBad or TrumpImmiPolicy & \\
      \hline
      K-Means 7 & Very messy. Links. & - \\ \hline
      K-Means 8 & European headlines and news. Some about the UK. & \\  
      & Send the ones that are relevant to UK policy themes & \\ \hline
      K-Means 9 & Detention, detention centers, solitary confinement as cruel. & DetainmentCruel \\ 
      \hline
 \end{tabular}}
        \caption{\textbf{First Iteration Immigration}: Patterns Identified in Initial Clusters and Resulting Themes}

    \label{tab:second_session_immi}
\end{table*}

\subsection{Topic Modeling Details}\label{app:topic_model_details}

To obtain LDA topics with Variational Bayes sampling we use the Gensim implementation~\cite{rehurek2011gensim}. To obtain LDA topics with Gibbs sampling we use the MALLET implementation~\cite{McCallumMALLET}. In both cases, we follow all the prepossessing steps suggested by \citet{NEURIPS2021_0f83556a}, with the addition of the words covid, vaccin* and immigra* to the list of stopwords. 

\subsection{Fine-Grained Results}\label{app:fine_grained_results}

\begin{figure}[H]
    \centering
    \includegraphics[width=\columnwidth]{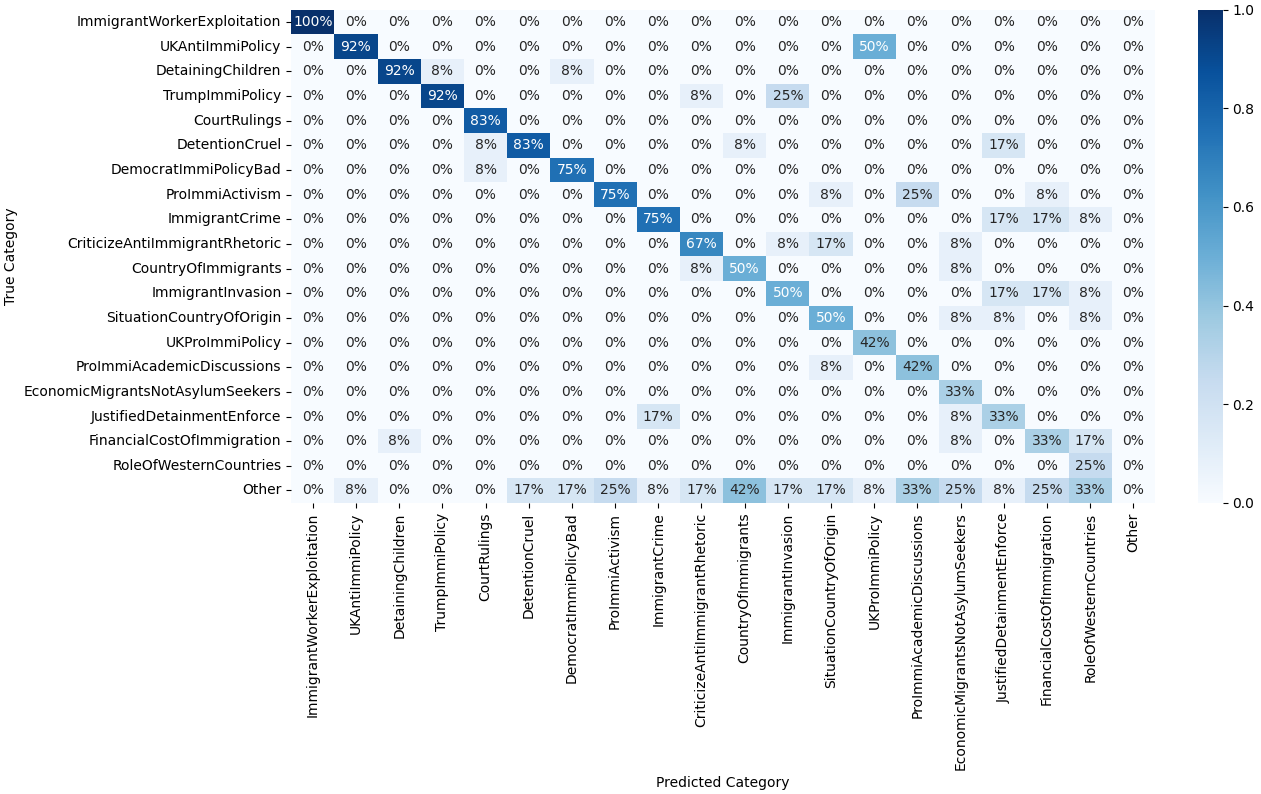}
    \caption{\textbf{Confusion matrix of Immigration themes after second iteration}. Values are normalized over the predicted themes (columns), and sorted from most accurate to least accurate.}
    \label{fig:cm_immi}
\end{figure}

\begin{figure}[H]
    \centering
    \begin{subfigure}{0.4\linewidth}
    \centering
    \includegraphics[width=\textwidth]{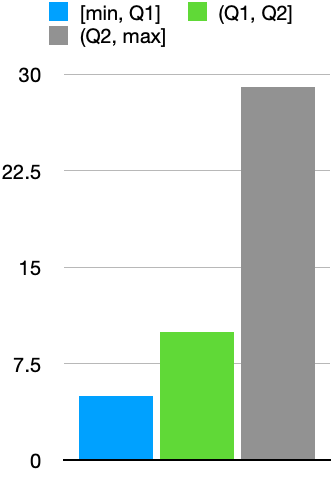}
    \caption{\textbf{Covid}}
    \end{subfigure}
    \begin{subfigure}{0.4\linewidth}
    \centering
    \includegraphics[width=\textwidth]{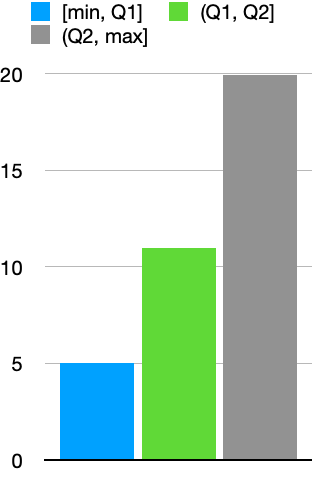}
    \caption{\textbf{Immigration}}
    \end{subfigure}
    \caption{Tweets that Do Not Match Current Set of Themes (True Category is ``Other'') at Different Intervals}
    \label{fig:other_distribution}
\end{figure}

The confusion matrix for Immigration can be seen in Fig.~\ref{fig:cm_immi}. Distribution of errors that do not match any existing theme, according to their similarity interval can be seen in Fig~\ref{fig:other_distribution}.

\subsection{Shifting Predictions between Iterations}\label{app:shifting_preds}

\begin{figure*}[t]
    \centering
    \includegraphics[width=\textwidth]{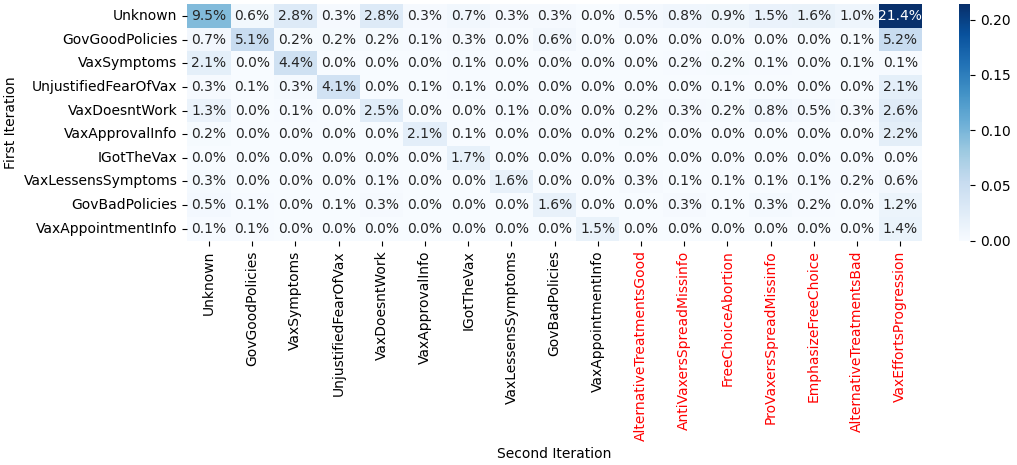}
    \caption{\textbf{Shifting predictions for Covid}. Themes added during second iteration are shown in red, and values are normalized over the full population.}
    \label{fig:shifting_preds_covid}
\end{figure*}

\begin{figure}[H]
    \centering
    \begin{subfigure}{0.4\linewidth}
    \centering
    \includegraphics[width=\textwidth]{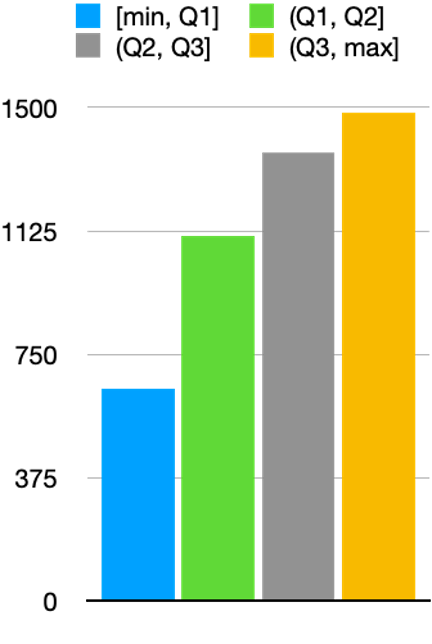}
    \caption{\textbf{Covid}}
    \end{subfigure}
    \begin{subfigure}{0.4\linewidth}
    \centering
    \includegraphics[width=\textwidth]{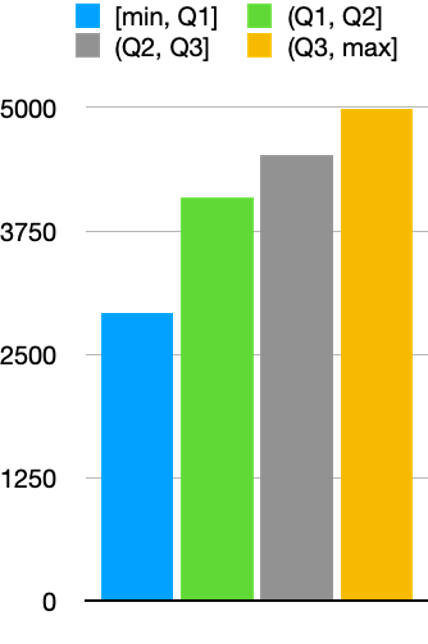}
    \caption{\textbf{Immigration}}
    \end{subfigure}
    \caption{Unmatched Predictions (Shifting from Named Theme to Unknown) at Different Intervals}
    \label{fig:unmatched_distributions}
\end{figure}

Heatmaps of shifting predictions for Covid can be seen in Fig. \ref{fig:shifting_preds_covid}. The distribution of the unmatched predictions for both Covid and Immigration, according to their similarity intervals can be seen in Fig. \ref{fig:unmatched_distributions}. Additionally, some examples of shifting predictions for the two themes with the most movement for the Immigration case can be seen in Tabs. \ref{tab:unmatched_western} and \ref{tab:unmatched_trump}.

\begin{table*}[t]
    \centering
    \scalebox{0.70}{\begin{tabular}{|>{\arraybackslash}m{2cm}|>{\arraybackslash}m{9cm}|>{\arraybackslash}m{9cm}|}
    \toprule
        Distance to Centroid & Example Tweets Kept on \textit{Role of Western Countries}  & Example Tweets Shifted to \textit{Unknown} \\
        \midrule
        0.27 & The U.S. Helped Destabilize Honduras. Now Honduran Migrants Are Fleeing Political and Economic Crisis &  Interesting that your problem is with "migrants", where the U.S. has issues with illegal aliens, that even our legal migrants wish to be rid of. \\
        \hline
        0.29 & These people are fleeing their countries DIRECTLY because of U.S. \# ForeignPolicy. If you don't like refugees. Don't create 'em. & The root causes of migration aren't being addressed ASAP, as they must be. The governments are all busy talking about stopping the consequences without  concrete plans to solve the causes. \\
        \hline
        0.30 & Don’t want migrants? Stop blowing their countries to pieces & What's missing in the US corporate news on migrants is the way American "aid" is used to overturn democracies, prop up strongmen and terrify the opposition.  \\
        \hline
    \end{tabular}}
    \caption{\textit{Role of Western Countries}: Examples of tweets kept on theme (Left) and shifted to unknown (Right) between the first and second iteration. On Right are the tweets closest to the theme centroid that shifted to \textit{Unknown}. On Left are tweets that did \textbf{\textit{not}} shift, but have the same distance.}
    \label{tab:unmatched_western}
\end{table*}

\begin{table*}[]
    \centering
    \scalebox{0.70}{\begin{tabular}{|>{\arraybackslash}m{2cm}|>{\arraybackslash}m{9cm}|>{\arraybackslash}m{9cm}|}
    \toprule
        Distance to Centroid & Example Tweets Kept on \textit{Trump Immigration Policy}  & Example Tweets Shifted to \textit{Unknown} \\
        \midrule
        0.24 & Racist \@realDonaldTrump wastes our tax money on locking up little kids in \#TrumpConcentrationCamps and steals from our military to waste money on his \#ReElectiomHateWall and spends little on anything else. & The anti-migrant cruelty of the Trump Admin knows no bounds. This targeting of migrant families is meant to induce fear and doesnt address our broken immigration system. We should be working to make our immigration system more humane, not dangerous and cruel.  \\
        \hline  
        0.25 & Trump promises immigration crackdown ahead of U.S. election &  This is unlawful and is directed at mothers with their children! He had no remorse for separating immigrants earlier, now he’s threatening their lives! It’s heart wrenching, but Trumpf has no heart! He’s void of feeling empathy! Read they are in prison camps? WH ignoring cries\\
        \hline
        0.26 & Trump to end asylum protections for most Central American migrants at US-Mexico border &  BBC News - Daca Dreamers: Trump vents anger on immigrant programme  \\
        \hline
    \end{tabular}}
    \caption{\textit{Trump Immigration Policy}: Examples of tweets kept on theme (Left) and shifted to unknown (Right) between the first and second iteration. On Right are the tweets closest to the theme centroid that shifted to \textit{Unknown}. On Left are tweets that did \textbf{\textit{not}} shift, but have the same distance.}
    \label{tab:unmatched_trump}
\end{table*}

\subsection{LDA vs. our Themes}\label{app:lda_themes}

\begin{figure}[ht]
    \centering
    \includegraphics[width=\columnwidth]{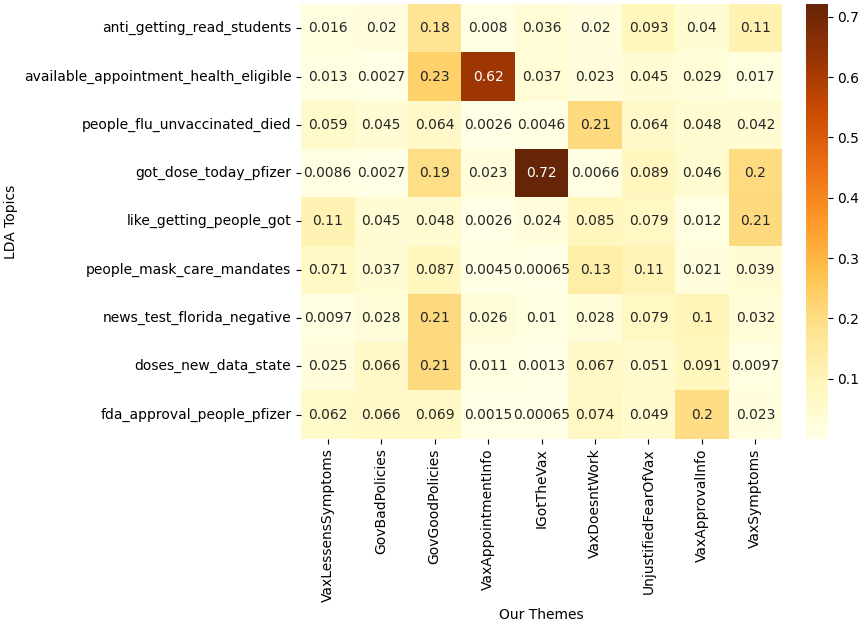}
    \caption{Overlap Coefficients between LDA Var. Bayes and our Themes (First Iteration for \textbf{Covid}). 
    }
    \label{fig:lda_theme_overlap_iter1_covid}
\end{figure}

\begin{figure*}[t]
    \centering
    \begin{subfigure}{2\columnwidth}
    \centering
    \includegraphics[width=\textwidth]{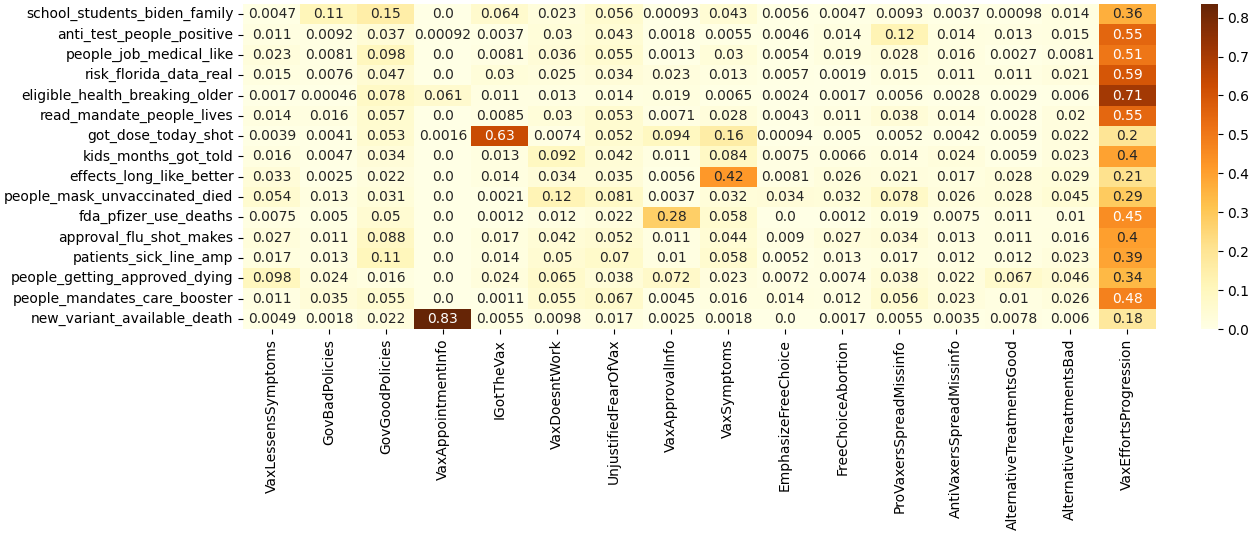}
    \caption{\textbf{Covid}}
    \end{subfigure}
    \begin{subfigure}{2\columnwidth}
    \centering
    \includegraphics[width=\textwidth]{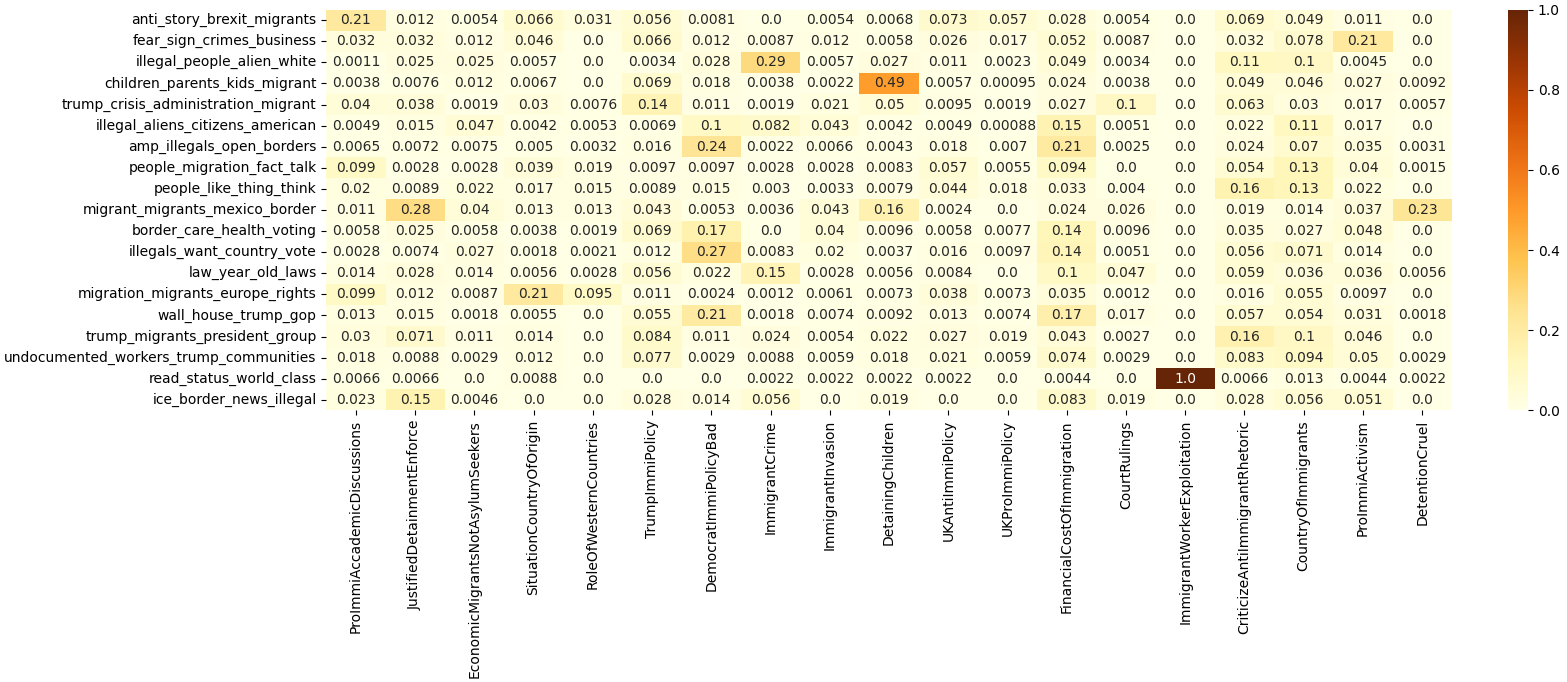}
    \caption{\textbf{Immigration}}
    \end{subfigure}
    \caption{Overlap Coefficients between LDA Var. Bayes and our Themes (Second Iteration).}
    \label{fig:lda_theme_overlap}
\end{figure*}

\begin{figure}[H]
    \centering
    \includegraphics[width=\columnwidth]{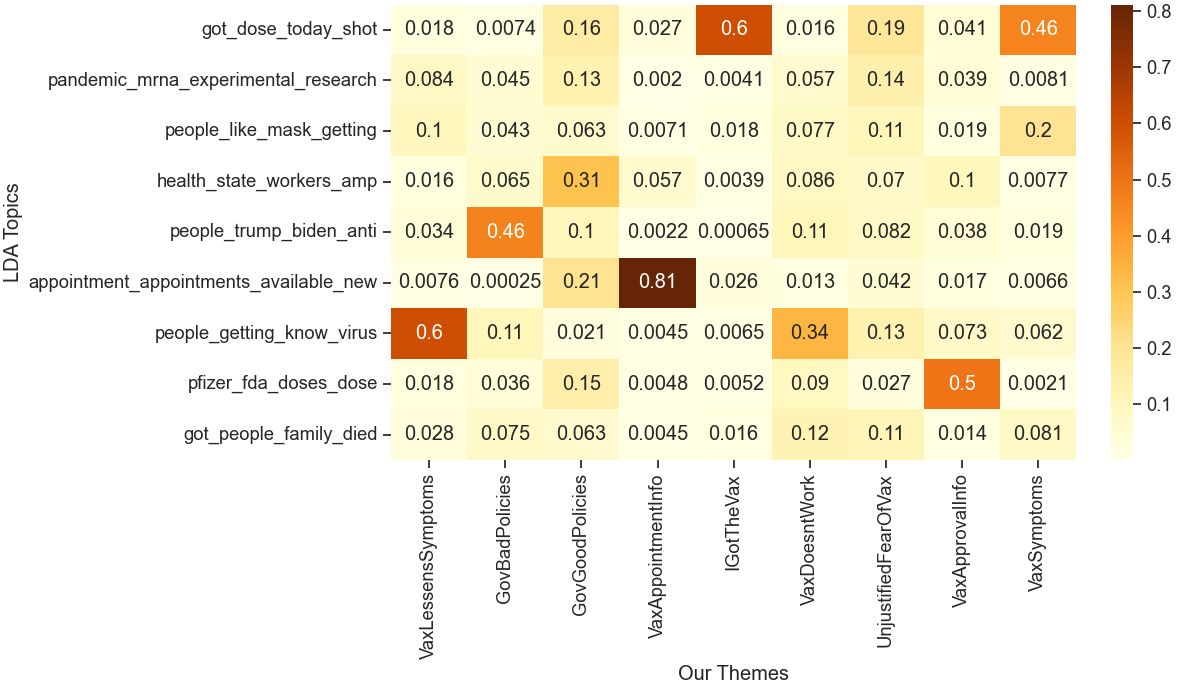}
    \caption{Overlap Coefficients between LDA Gibbs Sampling and our Themes (First Iteration for \textbf{Covid}). 
    }
    \label{fig:mallet_theme_overlap_iter1_covid}
\end{figure}

\begin{figure}[H]
    \centering
    \includegraphics[width=\columnwidth]{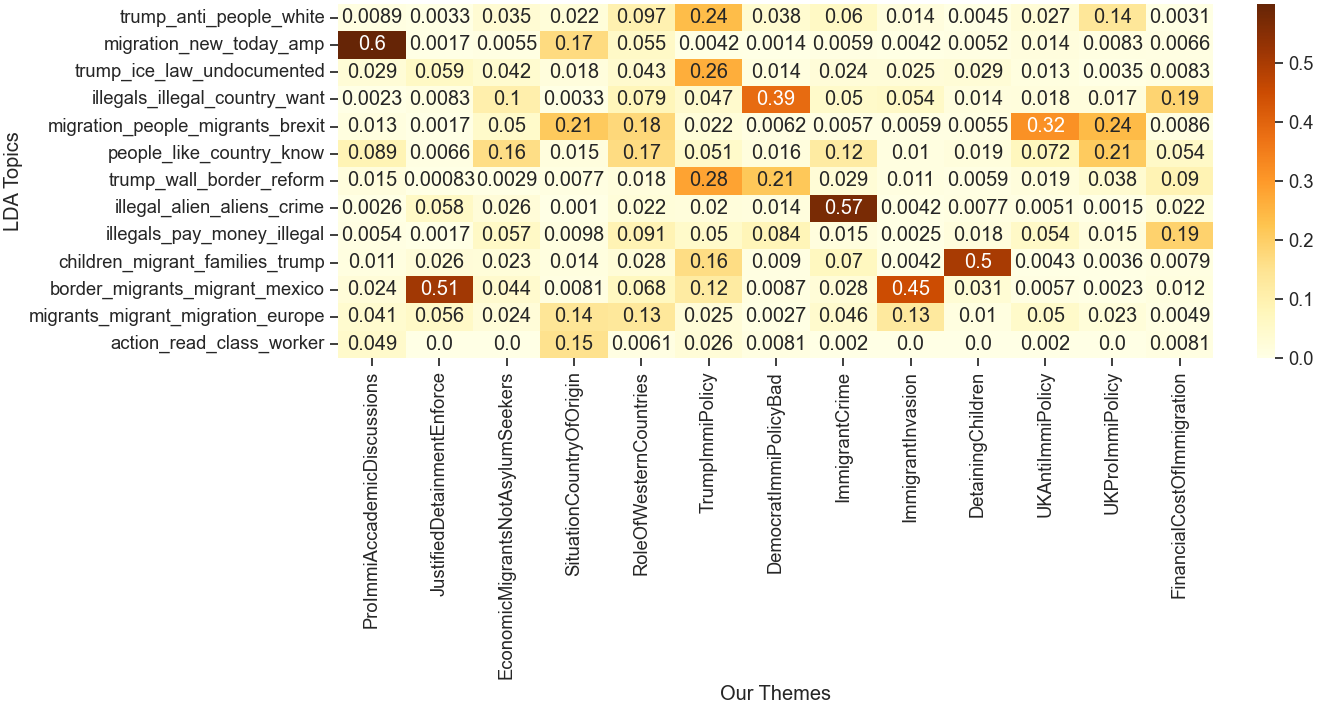}
    \caption{Overlap Coefficients between LDA Gibbs Sampling and our Themes (First Iteration for \textbf{Immigration}). 
    }
    \label{fig:mallet_theme_overlap_iter1_immi}
\end{figure}

\begin{figure*}[t]
    \centering
    \begin{subfigure}{2\columnwidth}
    \centering
    \includegraphics[width=\textwidth]{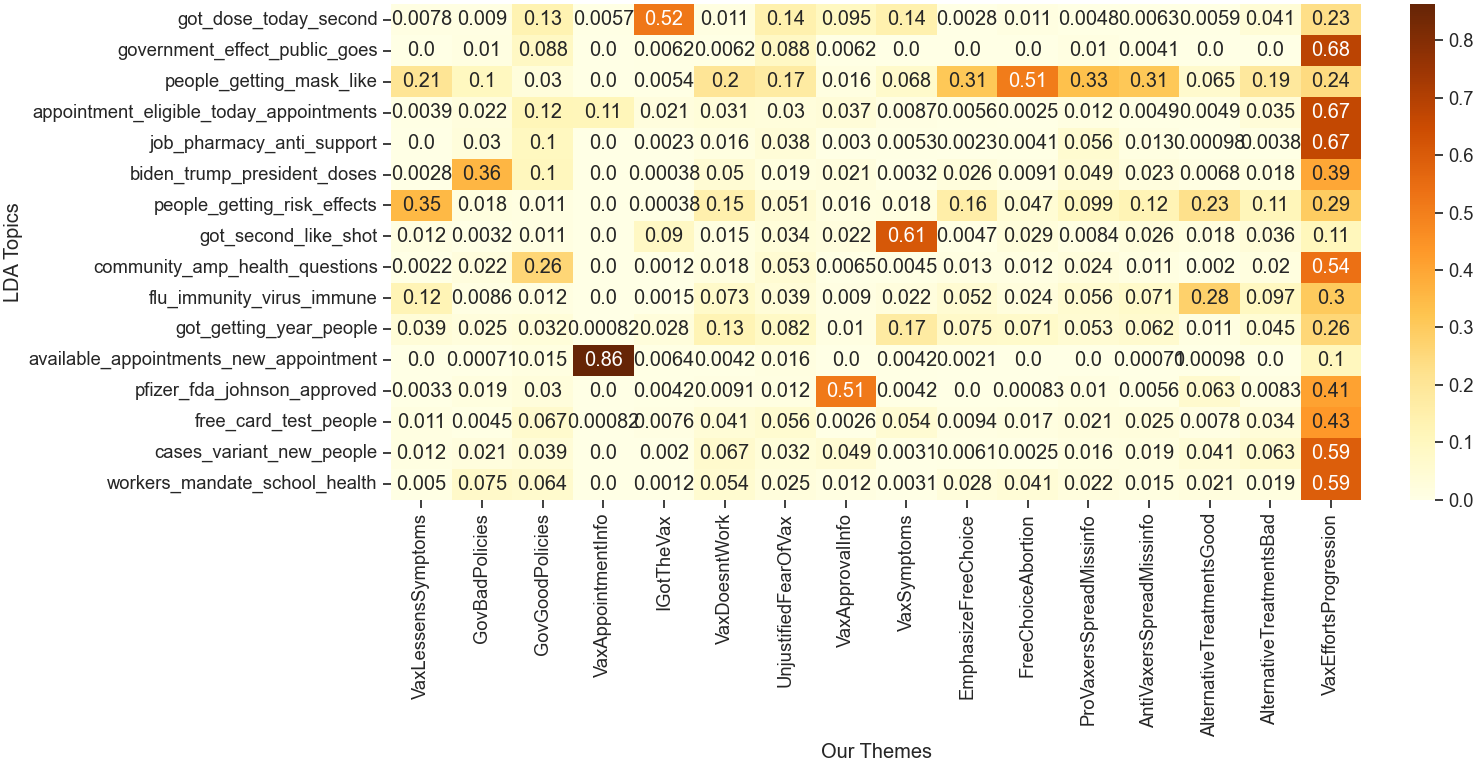}
    \caption{\textbf{Covid}}
    \end{subfigure}
    \begin{subfigure}{2\columnwidth}
    \centering
    \includegraphics[width=\textwidth]{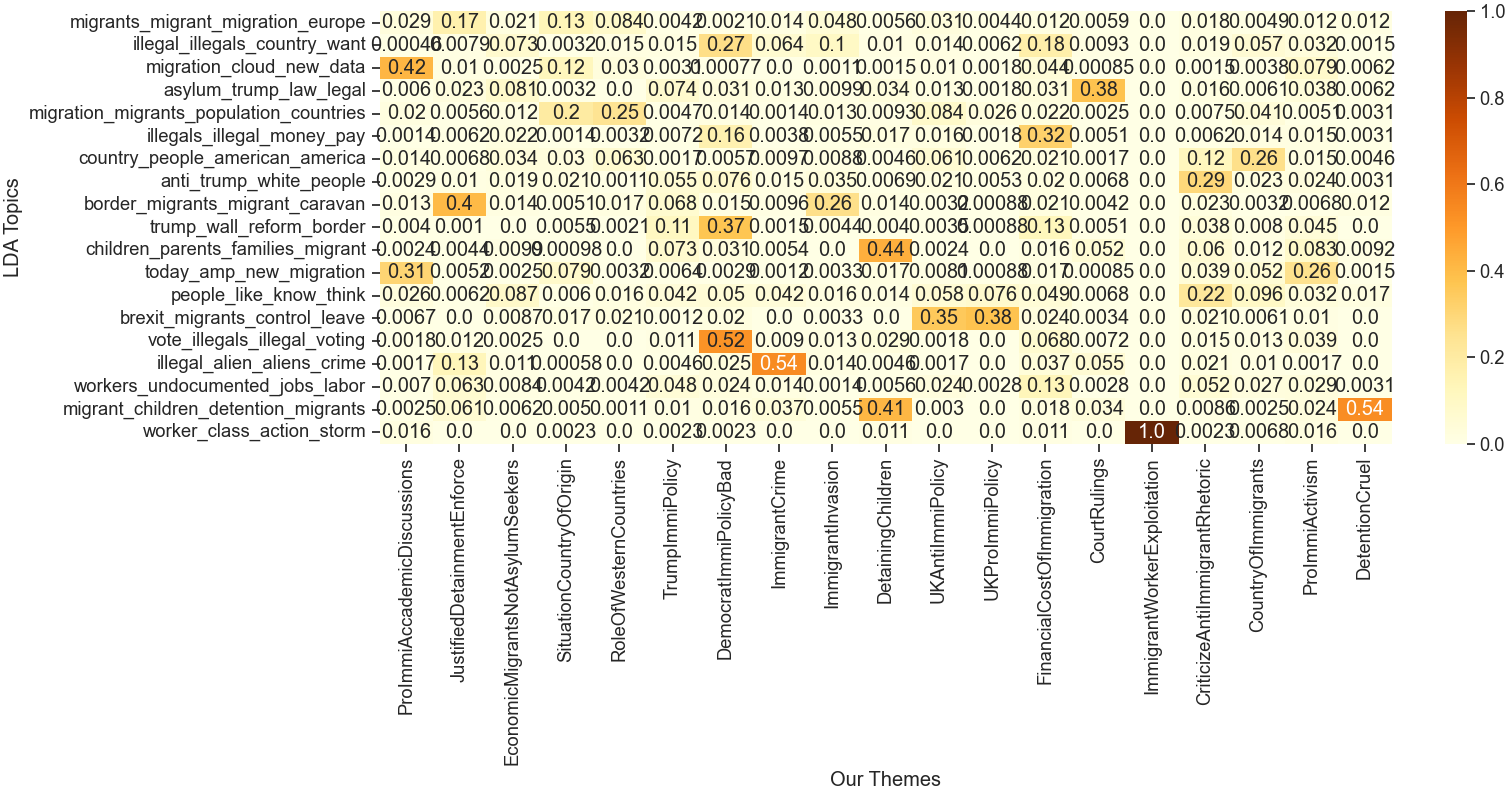}
    \caption{\textbf{Immigration}}
    \end{subfigure}
    \caption{Overlap Coefficients between LDA Gibbs Sampling and our Themes (Second Iteration).}
    \label{fig:mallet_theme_overlap}
\end{figure*}

An overlap coefficient heatmap between LDA topics with Variational Bayes sampling and our themes for the first iteration of Covid can be seen in Fig. \ref{fig:lda_theme_overlap_iter1_covid}. Similarly, they can be seen for the second iterations of both Covid and Immigration in Fig. \ref{fig:lda_theme_overlap}. We also include these heatmaps for LDA with Gibbs sampling in Figs. \ref{fig:mallet_theme_overlap_iter1_covid}, \ref{fig:mallet_theme_overlap_iter1_immi} and \ref{fig:mallet_theme_overlap}

\end{document}